# MR-LiDAR: A Multi-Resolution Roadside LiDAR Benchmark for Perception Diagnostics and Deployment Guidance

Shunlai Cui, Peng Cao*, Yuan Zhu, Yongjiang He, Jiacheng Yin, Xiao Huo, Gang Cao, and Xiaobo Liu

[1]***Abstract*—LiDAR configuration selection in roadside sensing remains constrained by limited controlled evidence across sensors. To address this gap, we present MR-LiDAR, a controlled cross-configuration benchmark comparing four LiDAR configurations from 16 to 128 beams on dynamic vehicle and vulnerable road user (VRU) targets. Its cross-configuration main set covers matched daytime, nighttime, and vehicle-occlusion conditions, while three deployment and operating condition subsets examine mounting height, target speed, and curved-road geometry. By reducing confounding from scene and trajectory variation, MR-LiDAR links sensor configuration to target point acquisition and algorithmic detection outcomes. We characterize distance-dependent point loss, define effective beam count as the number of vertical beams intersecting the target's angular extent, and introduce a point-based Occlusion Ratio. The results show that 16- and 32-beam configurations degrade earlier with distance, while VRUs exhibit greater point sparsity than motor vehicles. Nominal beam count alone is insufficient to explain roadside scanning capability. In the cross-configuration main set, the 80- and 128-beam configurations achieve comparable results, consistent with their nearly identical downward beam layouts. A uniform 80-beam counterfactual yields only about one third of the effective beam count of the actual non-uniform configuration. Illumination and the tested speed range have little influence on target-point-count patterns, whereas mounting height and near-field road geometry can substantially alter configuration ordering. Finally, an empirical minimal-sufficient configuration matrix combines frame-level geometric sufficiency with matched-box quality and includes threshold sensitivity analysis, providing bounded configuration guidance for the tested conditions. MR-LiDAR Benchmark is publicly available at https://github.com/ShunlaiC/MR-LiDAR-Benchmark.**

## I. Introduction

With the rapid deployment of Vehicle-to-Everything (V2X) infrastructure, roadside perception has become a critical component of cooperative automated driving systems [1]. LiDAR has been applied to point-cloud-based geometric measurement tasks [2], [3] and increasingly adopted in traffic perception applications. From a fixed and elevated viewpoint, roadside LiDAR complements on-board sensing by providing continuous coverage of the monitored road segment [4]. In large-scale deployment, LiDAR configuration selection requires matching sensing capability to specific application requirements. Since LiDAR detects objects by forming point clouds through laser scanning of object surfaces, target-point acquisition and downstream detection performance depend on LiDAR configurations, object categories, and target distances, as shown in Fig. 1. However, controlled empirical comparisons across LiDAR configurations remain limited, making it difficult to determine the minimum sufficient configuration for a given roadside perception task. A controlled real-world benchmark is therefore needed to isolate configuration-dependent effects, quantify cross-configuration differences, and provide evidence-based guidance for sensor selection and deployment.

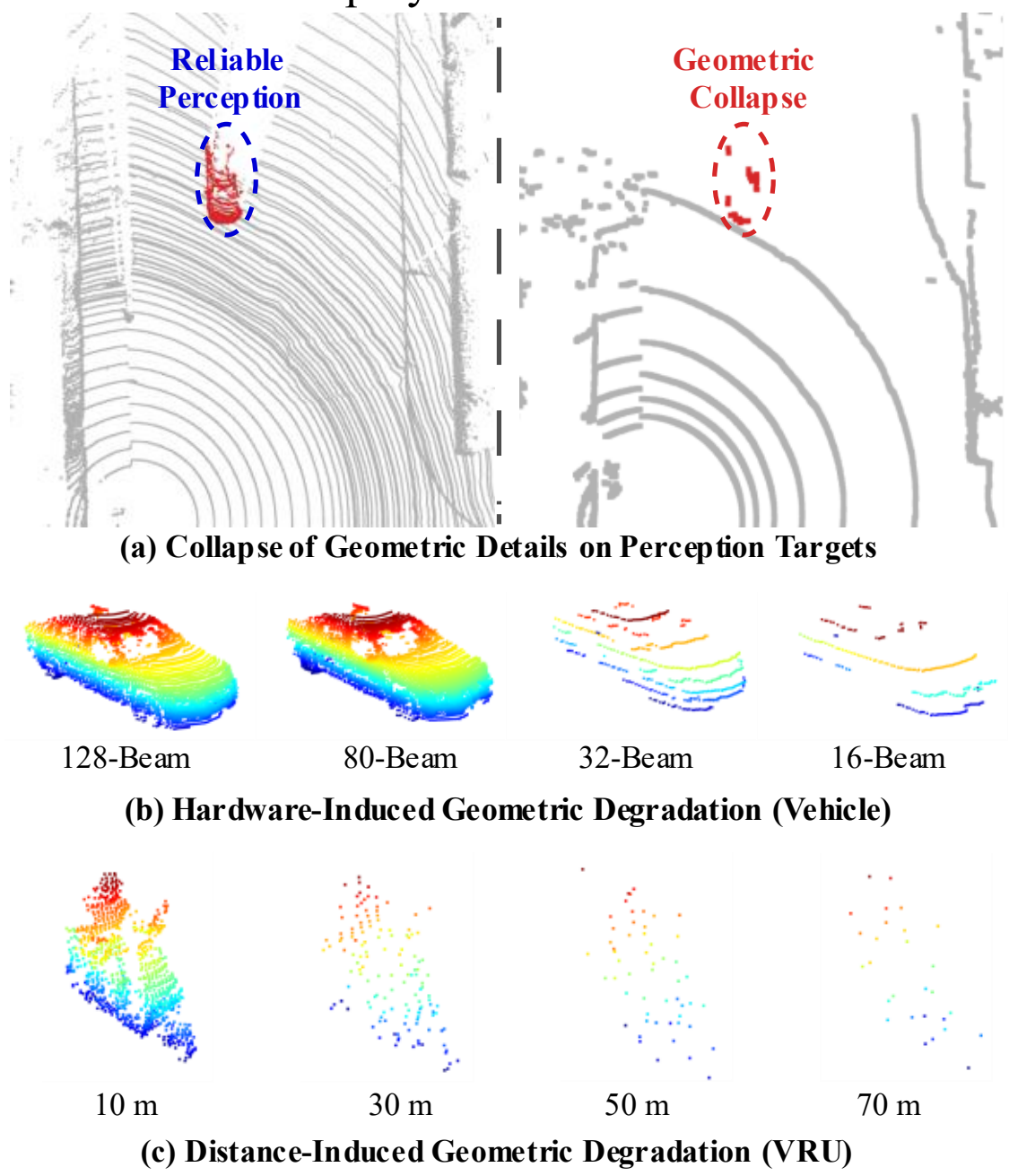

**Figure 1:** Geometric Collapse and Degradation in Roadside Perception.

[1]This work was supported by the National Natural Science Foundation of China under Grant No. 52572362 and the Fundamental Research Funds for the Central Universities No. 2682026CX060.

Shunlai Cui is with the School of Transportation and Logistics, Southwest Jiaotong University, Chengdu, 610031, China, and also with the Department of Electrical and Computer Engineering, Old Dominion University, Norfolk, VA, 23529, USA (e-mail: scui001@odu.edu).

Peng Cao, Xiao Huo, Gang Cao, and Xiaobo Liu are with the School of Transportation and Logistics, Southwest Jiaotong University, and also Intelligent Comprehensive Transportation Key Laboratory of Sichuan Province, Chengdu, 610031, China (e-mail: caopeng@swjtu.edu.cn; 2232418237@my.swjtu.edu.cn; cgandccy@my.swjtu.edu.cn; xiaobo.liu@swjtu.cn).

Yuan Zhu is with Dept. of Transportation Engineering, Transportation Institute, Inner Mongolia University, and also with the Inner Mongolia Engineering Research Center of Urban Transportation Data Science and Applications, Inner Mongolia University, Hohhot, 010020, China (e-mail: zhuyuan@imu.edu.cn).

Yongjiang He is with the School of Transportation and Logistics, Southwest Jiaotong University, Chengdu, 610031, China, and also with the Tangshan Institute, Southwest Jiaotong University, Tangshan, 063000, China (e-mail: yongjianghe@my.swjtu.edu.cn).

Jiacheng Yin is with the School of Automobile and Transportation, Xihua University, Chengdu, 610039, China (e-mail: yinjiacheng@xhu.edu.cn).

*Corresponding author: Peng Cao (e-mail: caopeng@swjtu.edu.cn).

Existing roadside LiDAR datasets, such as DAIR-V2X [5], Rope3D [6], and TUMTraf [7], are primarily designed for perception algorithm development and evaluation. Because LiDAR configuration is generally fixed within each acquisition setting, they do not directly support matched cross-configuration diagnostics in which sensor configuration is intentionally varied under controlled conditions. Nevertheless, they remain essential resources for large-scale algorithm evaluation and real-world perception research.

To address this gap, we present MR-LiDAR, a multi-resolution roadside LiDAR benchmark designed for controlled perception diagnostics and configuration evaluation. It compares four representative LiDAR configurations (16-, 32-, 80-, and 128-beam) covering both uniform and non-uniform vertical beam layouts, across multiple target categories including passenger cars, SUVs, and VRUs such as e-bikes and bicycles. By observing the same targets using different configurations under matched conditions, MR-LiDAR enables direct cross-configuration comparisons while reducing confounding effects caused by variations in scene geometry and target trajectories.

LiDAR selection is commonly posed as how many beams are required. Yet a beam contributes to a target only if its vertical angle falls within the angle the target subtends. Nominal beam count alone is therefore an incomplete descriptor of roadside scanning capability. In this study, we define the effective beam count as the number of vertical beams intersecting the target's vertical angular extent. Compared with nominal beam count, this metric more directly characterizes target-point acquisition and helps explain configuration-dependent differences in geometric detection quality. The main contributions are summarized as follows:

1) We establish MR-LiDAR, a controlled cross-configuration benchmark that compares four LiDAR configurations ranging from 16 to 128 beams using dynamic vehicle and VRU targets. The cross-configuration main set contains matched daytime, nighttime, and vehicle-occlusion sequences, while three deployment and operating condition subsets examine the effects of mounting height, target speed, and road geometry.
2) We conduct a series of empirical analyses to investigate how nominal beam count, vertical beam distribution, target distance, object category, vehicle occlusion, and acquisition geometry affect target-point acquisition and detection performance. The results demonstrate that nominal beam count alone is insufficient to characterize roadside LiDAR capability. Instead, effective beam count is more closely associated with target-point acquisition, matched-box overlap, and localization accuracy, while mounting height and road geometry can substantially alter the relative performance of different configurations in near field.
3) We establish an empirical reference guide for LiDAR configuration selection in roadside perception applications, including target-point-count analyses, cross-configuration detection performance comparisons, and a minimal-sufficient configuration matrix that jointly considers frame-level geometric sufficiency and matched-box quality. Threshold-sensitivity analysis and acquisition-condition sensitivity tests further evaluate the robustness of the matrix and define the scope of the resulting guidance.

## II. Related Work

### A. Roadside Perception Development & Benchmark Gaps

While foundational datasets like KITTI [8], nuScenes [9], and Waymo Open Dataset [10] have driven on-board perception research, ego-vehicle sensing is inherently affected by viewpoint-dependent occlusion and limited visibility [11] [12]. In this context, roadside perception can mitigate such fragmentation by providing a fixed infrastructure viewpoint and enabling continuous trajectory capture [13]. Simulation-based roadside datasets like V2X-Sim [14] support collaborative perception research but suffer from the persistent sim-to-real domain gap [15]. Shifting to real-world data, large-scale benchmarks such as DAIR-V2X [5] and TUMTraf [7] have set new standards for environmental diversity and annotation volume. However, the use of a single fixed LiDAR configuration in each of these datasets makes it impossible to conduct cross-tier LiDAR comparisons [16]. Consequently, controlled empirical evidence for identifying minimal-sufficient LiDAR configurations under matched roadside conditions remains limited.

### B. LiDAR Perception Diagnostics

Beyond single-configuration algorithm benchmarks, multi-LiDAR resources such as LIBRE [16] and Schulte-Tigges et al. [17] support hardware characterization across multiple sensor configurations, including controlled target measurements of properties such as range accuracy and point density. These studies demonstrate the value of multi-configuration hardware analysis. However, they do not provide repeated matched roadside target trajectories in which LiDAR configuration is intentionally varied as the primary experimental factor for dynamic perception diagnostics.

Complementing these hardware-characterization studies, research on LiDAR design and deployment geometry has examined how beam allocation and LiDAR placement affect task-relevant sensing. Zhuang et al. [18] showed that task-specific vertical beam allocation can improve detection range at a fixed beam budget. Recent roadside studies have further considered beam distribution jointly with sensor placement [4], [19] and with dynamic vehicle occlusion [20]. These works motivate empirical analysis of how vertical beam layouts affect target point acquisition and detection outcomes under matched roadside trajectories.

As summarized in Table I, the dataset comparison distinguishes the presence of multiple LiDAR configurations from the role that LiDAR hardware plays in the dataset design. In this study, controlled dynamic hardware diagnostics denotes a dataset design that combines: (i) multiple LiDAR configurations, (ii) dynamic traffic targets, and (iii) matched or otherwise controlled acquisition conditions in which LiDAR configuration is intentionally treated as an experimental variable. Accordingly, LIBRE [16] and Schulte-Tigges et al. [17] provide resources for hardware characterization, whereas MR-LiDAR is designed for controlled cross-configuration diagnostics with dynamic roadside targets.

TABLE I
COMPARISON OF MR-LIDAR WITH REPRESENTATIVE LIDAR DATASETS

| Dataset | Viewpoint | LiDAR Types | Target Setting | Hardware-Diagnostic Design | Primary Focus |
|---|---|---|---|---|---|
| KITTI [8] | Ego-Vehicle | 1 × 64-beam | Dynamic | Not designed | On-board Algorithm Benchmarking |
| nuScenes [9] | Ego-Vehicle | 1 × 32-beam | Dynamic | Not designed | On-board Multimodal Fusion |
| LIBRE [16] | Ego-Vehicle | 10 Types | Static | Static-only hardware comparison | Static Hardware Comparison |
| Schulte-Tigges [17] | Ego-Vehicle | 6 Types | Static | Static-only hardware comparison | Solid-State LiDAR Comparison |
| DAIR-V2X-I [5] | Roadside | 1 × 300-beam | Dynamic | Not designed | Cooperative Perception |
| TUMTraf [7] | Roadside | 1 × 64-beam | Dynamic | Not designed | Roadside Algorithm Benchmarking |
| Int2Sec [21] | Roadside | 2 × 80-beam | Dynamic | Not designed | Roadside Digital Twin |
| **MR-LiDAR (Ours)** | **Roadside** | **16/32/80/128-beam** | **Dynamic** | **Controlled dynamic cross-configuration diagnostics** | **Dynamic Hardware Diagnostics** |

## III. THE MR-LIDAR BENCHMARK: DESIGN AND STATISTICS

### A. Experimental Design & Data Collection

To investigate how different factors affect the perception capability of roadside LiDAR, we organized the data collection into two components: (i) a **cross-configuration main set** for controlled perception diagnostics and (ii) **three deployment and operating condition subsets** for examining the effects of road geometry, mounting height, and target speed. In the cross-configuration main set, LiDAR configuration was treated as the primary experimental factor, while object category, illumination condition, and occlusion state were varied under matched acquisition conditions. The three subsets each vary one additional deployment or operating factor while maintaining the relevant sensor configuration, target characteristics, and processing settings. This organization reduces scene- and trajectory-related confounding in the cross-configuration comparisons and provides bounded evidence of whether the principal geometric trends persist under different deployment and operating conditions.

For the cross-configuration main set, we selected a 200-meter straight road segment on the campus of Southwest Jiaotong University, as shown in Fig. 2. The straight-road setting was chosen to minimize the effects of road slope and curvature on cross-configuration comparisons. We employed four LiDARs to establish the MR-LiDAR benchmark. As detailed in Table II, these sensors span from entry-level models (RS-LiDAR-16 and RS-Helios-1615) to high-end models (RS-Ruby-80V and RS-Ruby-Plus). Each LiDAR was mounted sequentially on the same standardized 1.5-meter bracket, thereby maintaining a consistent roadside viewpoint.

The targets include passenger car, SUV, e-bike, and bicycle. To ensure coverage of both motorized vehicles and VRUs, we conducted repeated data acquisition trials for each object type under controlled conditions. During each trial, the target traversed a continuous trajectory from the western end (<-80m) of the road segment to the eastern end (>80m) at a controlled speed of 10 km/h. This low-speed setting obtains a high frame density per meter, enabling fine-grained recording of the target's geometric evolution throughout the entire traversal process.

For illumination condition, data in the cross-configuration main set were collected under both sunny daytime (28,256 Lux) and low-light nighttime (81 Lux). For the occluded condition, we designed a dynamic occlusion scenario where the target vehicle travels behind a stationary obstacle vehicle located at $y$ = 1.25 m, $x$ = 5 m in the LiDAR coordinate system. This setup creates a full occlusion process, during which the target transitions from non-occluded to partially or fully occluded, and then reappears. Moreover, a Mavic 3 Pro drone was utilized to record aerial video following the collection configuration described in [22]. This bird's eye view (BEV) footage provides auxiliary visual reference and facilitates precise annotation even when the target is completely invisible to the LiDAR.

In addition to the cross-configuration main set, we collected three deployment and operating condition subsets:

1) **Mounting-height Subset:** To evaluate the influence of roadside LiDAR installation height, we collected paired Car sequences at mounting heights of 1.5 and 2.5 m on the straight-road segment. The comparison covered all four LiDAR configurations under daytime conditions at a target speed of 30 km/h. The road segment, target category, trajectory, and annotation procedure were kept consistent across the two height conditions. As analyzed in Section IV-C, this subset was used to examine height-dependent changes in near-field point density, theoretical blind-zone radius, and the ground-return occlusion shadow behind the target.

2) **Target-speed Subset:** To determine whether the 10 km/h target speed used in the cross-configuration main set introduced speed-dependent bias, we compared 128-beam Car sequences collected at 10, 30, and 50 km/h. All sequences were acquired on the same straight-road segment at a mounting height of 1.5 m under daytime conditions and were processed using the same annotation and preprocessing procedures. To isolate speed-dependent effects while reducing the influence of point sparsity, the analysis in Section IV-D focuses on the 128-beam Car sequences at 10, 30, and 50 km/h. The high target-point density of the 128-beam configuration facilitates the examination of speed-related changes, including scan-induced geometric distortion and reduced trajectory sampling density. The results are presented in Section IV-D.

3) **Curved-road Subset**: To determine whether the cross-configuration point-count patterns observed on the straight-road segment persisted under road curvature, we collected an additional curved-road subset. This subset used the same four LiDAR configurations and annotation procedure as the cross-configuration main set, with a target speed of 10 km/h. It included Car and Bicycle targets acquired at mounting heights of 1.5 and 2.5 m. For each LiDAR–target–height combination, one left-turn and one right-turn traversal were recorded, yielding 32 sequences and 7,288 annotated frames. As analyzed in Section IV-E, this subset was used to evaluate the sensitivity of the straight-road point-count patterns to road curvature.

These subsets were designed to examine whether the principal findings remain stable under changes in key deployment and operating factors, including height, target speed, and road geometry. Accordingly, the cross-configuration main set supports controlled cross-configuration diagnostics, while the subsets provide bounded evidence regarding the robustness of the observed trends and their dependence on specific acquisition conditions.

TABLE II
DETAILED CONFIGURATIONS OF THE LIDAR HARDWARE

| LiDARs | RS -LiDAR-16 | RS-Helios-1615 | RS-Ruby-80V | RS-Ruby-Plus |
|---|---|---|---|---|
| **Number of Beams** | 16 | 32 | 80 | 128 |
| **Beam Distribution** | Uniform | Uniform | Non-uniform | Non-uniform |
| **Horizontal FOV** | 360° | 360° | 360° | 360° |
| **Horizontal Resolution** | 0.1°-0.4° | 0.2°-0.4° | 0.1°-0.4° | 0.1°-0.4° |
| **Vertical FOV** | ±15° | -16°~+15° | -25°~+0.2° | -25°~+15° |
| **Vertical Resolution** | 2° | 1° | Up to 0.1° | Up to 0.1° |

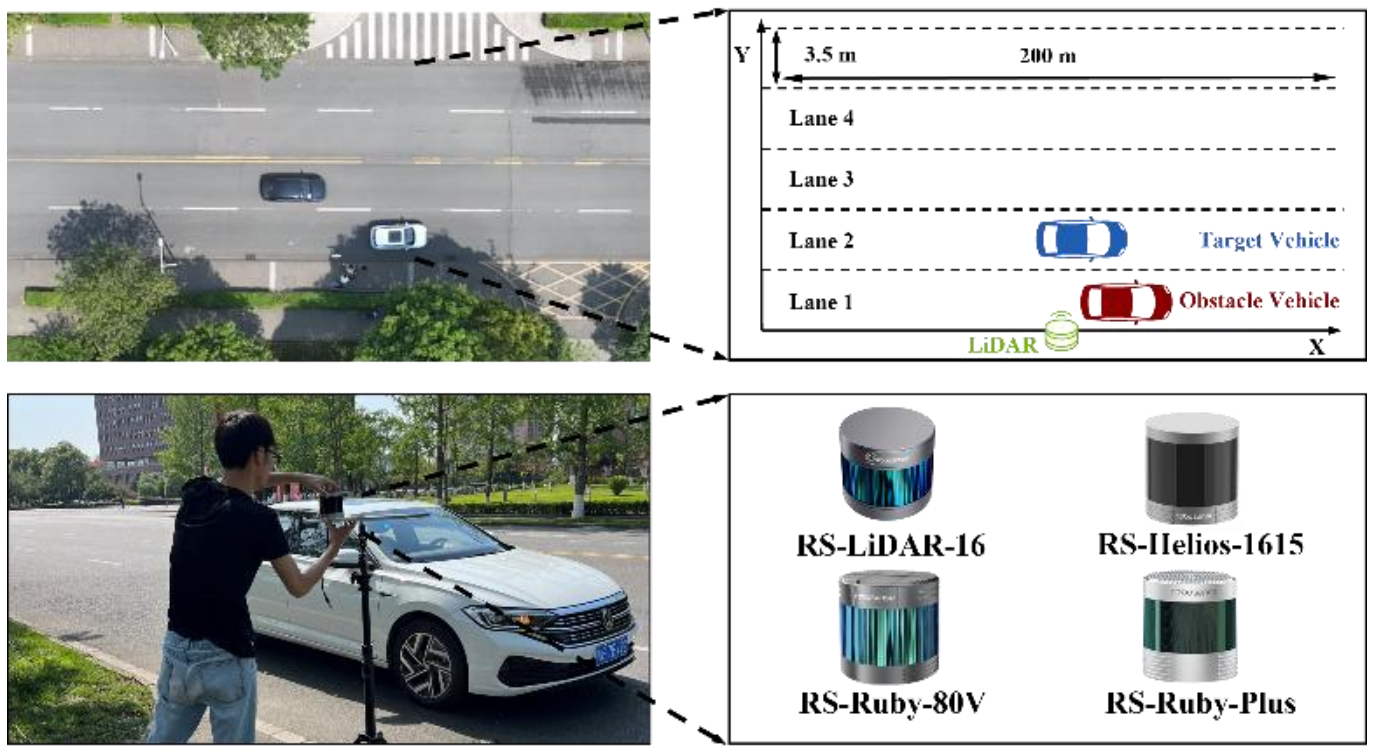

**Figure 2:** Controlled Experimental Setup and LiDAR Configuration.

*B. Data Preprocessing and Labeling*

To ensure annotation quality and consistency, we used SUSTechPOINTS [23] for semi-automatic annotation. For each target sequence, accurate 3D bounding boxes were first manually created in representative frames in which the target was clearly visible. Using the platform's multiple-instance editing function in the "Auto (no rotation)" mode, these manually initialized boxes were propagated to the remaining frames through temporal interpolation and motion-based extrapolation. All propagated boxes were then visually inspected and manually corrected. Vehicle dimensions were standardized according to official manufacturer specifications, while box centers and orientations were manually verified against the corresponding point clouds. For occluded vehicle frames, adjacent visible frames and drone video were used as additional references. This procedure produced quality-controlled 3D bounding-box annotations together with their associated object point clouds.

To assess the consistency of this annotation procedure across sensors, we quantified cross-LiDAR annotation agreement using four independently annotated Car sequences. Since the four LiDAR configurations share a common roadside coordinate system and the target vehicle traverses the same road segment, annotated box centers from different LiDAR configurations can be compared at matched longitudinal positions $X$. Within the common $X$-range covered by all four sequences, box centers were resampled onto a shared grid, and the center distance was computed for all six LiDAR pairs at each grid point. Pooled across all pairs, the center distance $d_c$ has a median of 0.18 m and a mean of 0.20 m, equivalent to approximately 4% of the vehicle length. Moreover, the disagreement did not increase monotonically as nominal beam count decreased: the difference between pairs involving the 16- or 32-beam LiDAR and the 80/128-beam pair was only approximately 1.4% of the vehicle length, as shown in Fig. 3. These results demonstrate that the independently annotated boxes remain consistent across LiDAR configurations, with no systematic degradation in annotation agreement for lower-beam sensors.

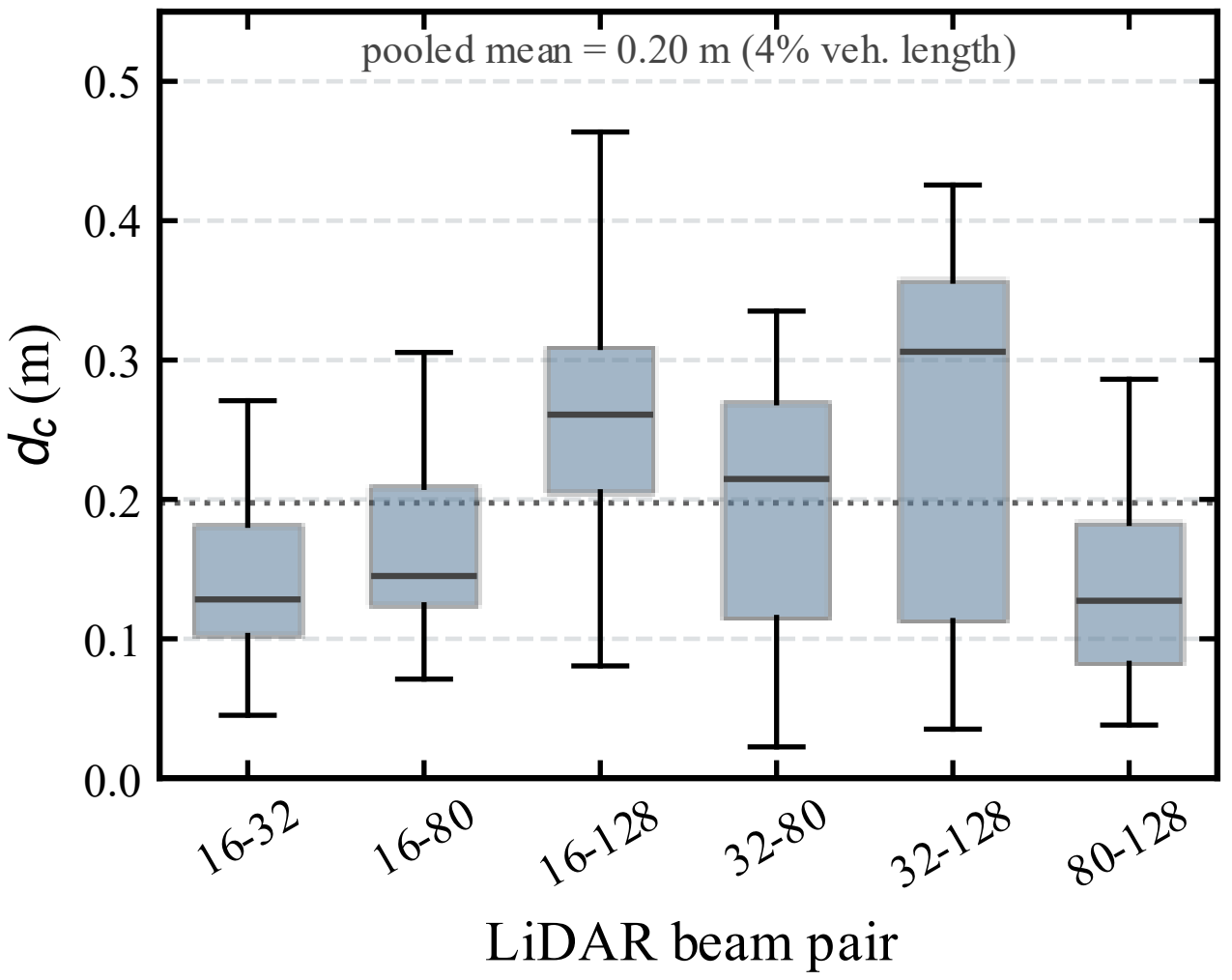

**Figure 3:** Cross-LiDAR annotation consistency. Center distance $d_c$ between annotated box centers for each of the six LiDAR pairs, evaluated at matched longitudinal positions on a shared resampling grid using four independently annotated Car sequences.

To ensure a consistent annotation process, a valid target frame is defined as one in which the target is confirmed through temporally continuous observations and its refined candidate 3D bounding box contains at least 40 LiDAR points. This threshold was adopted as a uniform empirical operational criterion based on frame-by-frame observations of targets. Below 40 points, a target generally appeared as a sparse moving cluster, and its spatial extent, position, and orientation could not be stably determined from a single LiDAR frame under our annotation process. Above 40 points, sufficient local target geometry became visually distinguishable for stable annotation, typically requiring only fine adjustment of the box position and orientation. Candidate frames with fewer than 40 LiDAR points inside the refined 3D box were therefore removed from the final annotated benchmark and standard detection evaluation.

### C. Vertical Beam Distribution Characterization

Across the four LiDAR configurations, beam count and vertical beam distribution are treated as the primary hardware attributes for diagnostics. Accordingly, we characterize the vertical beam layouts of the four LiDARs from the acquired data, so that the reported geometry corresponds to the LiDARs as deployed. For each beam, the vertical angles of all returns assigned to that beam were computed across frames, and their median was taken as a robust estimate of the beam angle, as shown in Table III.

The 16- and 32-beam LiDARs have uniform beam distributions, with median adjacent spacings of 1.98° and 0.99°. The 80- and 128-beam LiDARs have non-uniform beam distributions, both with a median adjacent spacing of 0.100° and a maximum adjacent spacing greater than 5.3°, as illustrated in Fig. 4. The 80-beam LiDAR places 96.3% of its beams below the horizon, against 60.2% for the 128-beam LiDAR. The two high-beam LiDARs both provide 77 beams below the horizon, and after sorting, the median absolute difference between their downward beam angles is 0.00007°, with a maximum of 0.070°. Over the downward angular range relevant to road targets, their vertical beam layouts are therefore nearly identical. The 48 additional beams of the 128-beam LiDAR lie above the horizon, where it carries 51 upward beams against 3 for the 80-beam LiDAR.

TABLE III
MEASURED VERTICAL BEAM PROFILES OF THE FOUR LIDAR CONFIGURATIONS

| LiDAR configuration | Beam count | Vertical angular range | Downward beam count | Median Gap | Mean Gap |
|---|---|---|---|---|---|
| **16-beam** | 16 | −14.8° to +14.9° | 8 | 1.98° | 1.98° |
| **32-beam** | 32 | −15.8° to +14.9° | 17 | 0.99° | 0.99° |
| **80-beam** | 80 | −24.8° to +0.2° | 77 | 0.10° | 0.31° |
| **128-beam** | 128 | −24.8° to +15.0° | 77 | 0.10° | 0.31° |

To relate these patterns to LiDAR measurements on a target, we define the effective beam count $k_{eff}$ as the number of LiDAR beams whose measured vertical angles fall within the vertical angular interval subtended by the target at a given range. It is obtained from the beam angles of Table III and the annotated box dimensions. Unlike nominal beam count, $k_{eff}$ depends jointly on beam layout, target geometry, and acquisition geometry.

Because the physical LiDARs differ in both beam count and vertical beam distribution, direct comparison of their measured profiles cannot isolate the contribution of beam distribution. To quantify this effect while holding nominal beam count and vertical field of view constant, we constructed a counterfactual LiDAR configuration having the same beam count of 80 and the same vertical field of view as the RS-Ruby-80V. For a Car target, the actual RS-Ruby-80V which has a non-uniform beam distribution, yields $k_{eff}$ of 43, 21, and 10 at 20 m, 40 m, and 80 m, respectively. Whereas the uniform 80-beam baseline yields 14, 7, and 4. The non-uniform beam distribution therefore provides 3.07×, 3.00×, and 2.50× the effective beam counts of the uniform distribution, respectively. The non-uniform configuration likewise yields a higher $k_{eff}$ for the other three target classes across the measured range. Thus, concentrating a fixed beam count into the depression-angle band occupied by road targets can increase target-relevant scanning by approximately threefold without changing the nominal beam count or vertical field of view.

These measurements provide a geometric basis for the cross-configuration comparisons developed in Sections IV and V. The comparable point-count and detection performance of the 80- and 128-beam sensors reported in these sections is consistent with their nearly identical downward beam layouts. Accordingly, the effective beam count $k_{eff}$ provides a more direct descriptor of target-relevant vertical scanning than nominal beam count alone.

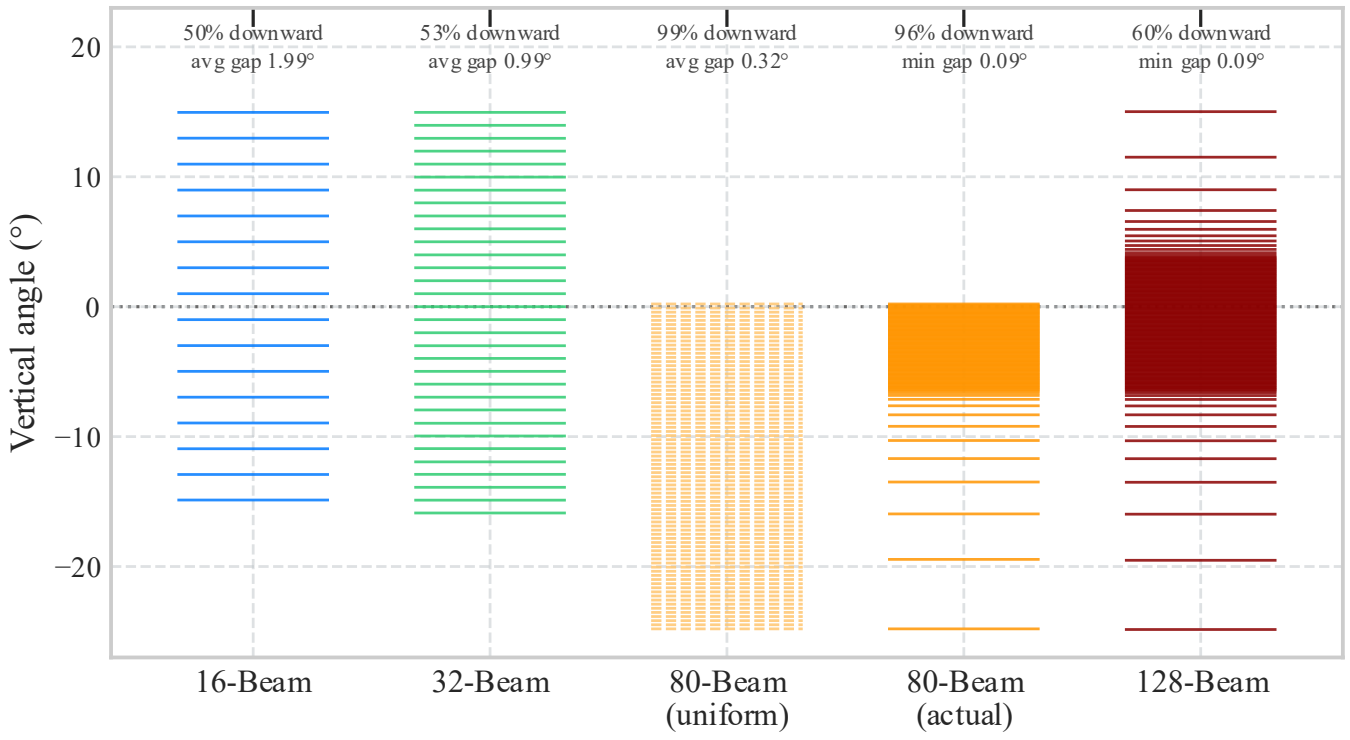

**Figure 4:** Measured vertical beam-angle distribution of the four LiDAR configurations.

### D. Data Structure

To facilitate the use of the collected dataset, we adopt a target-centric organization, in which each data sequence captures the complete trajectory of a single target. As illustrated in Fig. 5, the dataset is hierarchically structured according to *Illumination Condition, LiDAR Type, Object Class, and Sequence Index*. This design supports matched cross-configuration comparisons, such as comparisons between the 16- and 128-beam sensors, under highly consistent target trajectories.

Each sequence contains four components: (1) the full-scene LiDAR point cloud (.pcd); (2) 3D bounding box annotations (.json); (3) pre-cropped target point clouds (vehicle_pcd), which isolate the object geometry to simplify density analysis without manual cropping; and (4) the corresponding drone-captured aerial video.

```
MR-LiDAR_Benchmark/
├── 20250409-Day/                  <-- Level 1: Scenario (Illumination)
│   ├── 128-Beam/                  <-- Level 2: Sensor
│   │   ├── Car/                   <-- Level 3: Class
│   │   │   └── Seq_20250409.../   <-- Level 4: Sequence
│   │   │       ├── lidar/
│   │   │       │   ├── 0001.pcd
│   │   │       │   └── ...
│   │   │       │   ├── label/
│   │   │       │   ├── 0001.json
│   │   │       │   └── ...
│   │   │       ├── vehicle_pcd/
│   │   │       │   ├── 0001.pcd
│   │   │       │   └── ...
│   │   │       └── video/
│   │   │           └── drone_view.mp4
│   │   └── Car-Occlusion/
│   │   └── SUV/
│   │   └── E-bike/
│   │   └── Bicycle/
│   └── 80-Beam/
│   └── 32-Beam/
│   └── 16-Beam/
└── 20250409-Night/
└── 20250324-Day/
└── 20250324-Night/
```

**Figure 5:** Data structure of the cross-configuration main set.

*E. General Statistics*

The cross-configuration main set contains 10,497 frames, each annotated with a high-precision 3D bounding box. These frames are organized into 32 controlled sequences, each lasting approximately 60 seconds and capturing the complete approach-and-departure trajectory of a target.

Rather than maximizing scene diversity or training volume, the cross-configuration main set prioritizes matched observations across LiDAR configurations under controlled target and acquisition conditions, while covering multiple target classes, distance ranges, illumination conditions, and occlusion states. This design reduces confounding from uncontrolled scene variation. MR-LiDAR is therefore intended to complement, rather than replace, large-scale datasets such as DAIR-V2X and TUMTraf, which support perception-model development through broader scene coverage and larger data volumes. The controlled scale of the cross-configuration main set consequently defines the scope of the cross-configuration comparison rather than claiming broad urban-scene coverage.

Fig. 6(a) shows the distribution of data frames across different traffic participants. Passenger cars (52.0%) and SUVs (22.8%) constitute the motorized vehicle category, while VRUs, including e-bikes and bicycles, collectively account for 25.2% of the data. In addition, the cross-configuration main set contains 5,454 vehicle frames, of which 2,612 involve occlusion and are denoted as Car-Occ, capturing the occlusion dynamics of a single moving target passing behind a stationary obstacle vehicle.

Fig. 6(b) shows the overall distribution of data frames between daytime and nighttime conditions. Daytime and nighttime data account for 64.3% and 35.7% of the cross-configuration main set, respectively. A more detailed breakdown across the experimental variables is provided in Table IV. The nighttime proportions are relatively consistent across the four LiDAR configurations, ranging from 35.0% to 39.0%. Similarly, for target distances, the nighttime proportion remains between 34.4% and 36.4%. The occluded-vehicle data also contain nighttime frames, accounting for 51.3% of all occluded frames. The class distribution is not fully balanced between illumination conditions. Nighttime data include 2,669 Car frames and 1,076 SUV frames, whereas E-bike and Bicycle sequences were collected only during daytime.

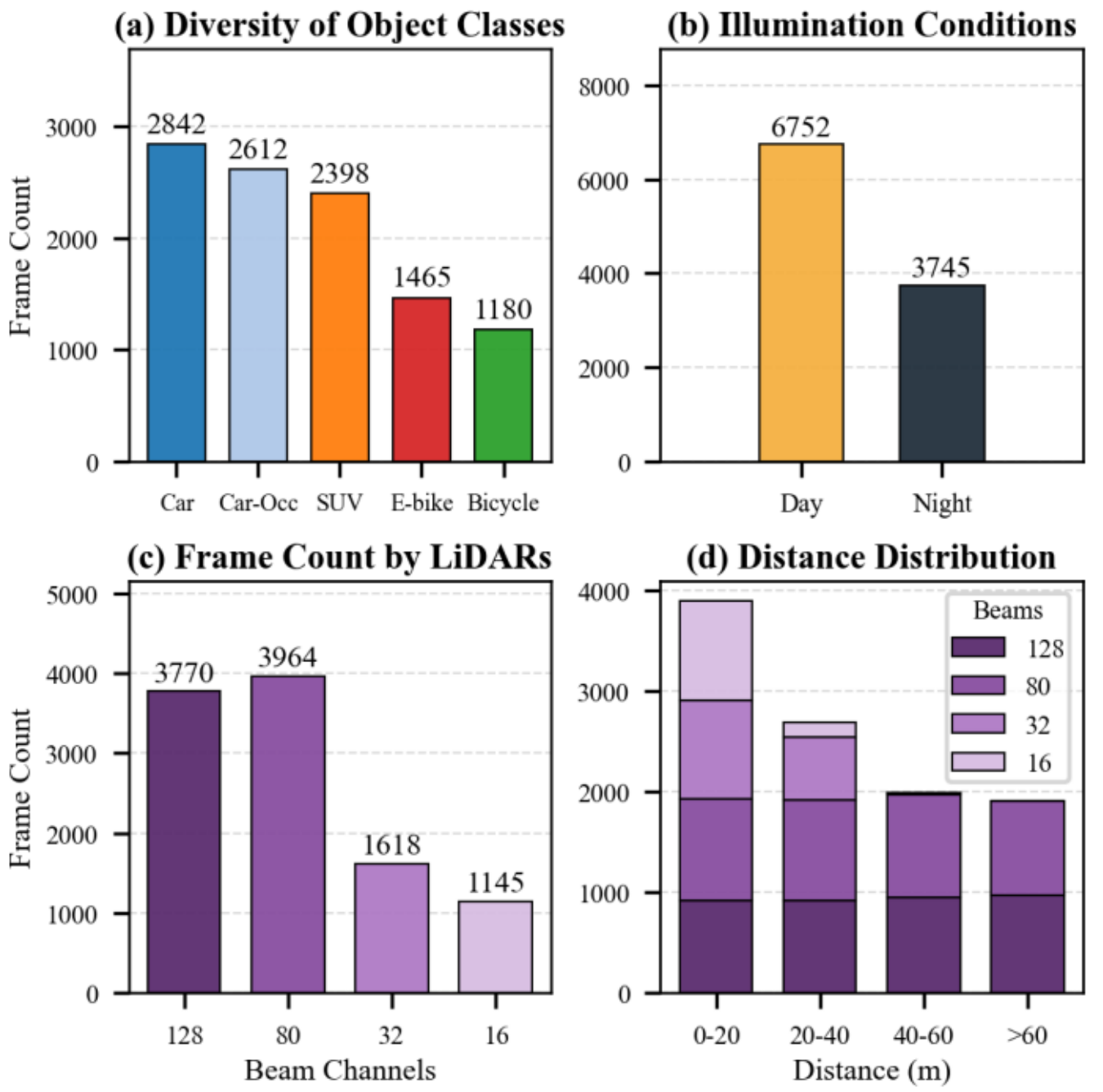

**Figure 6:** Statistical Distribution of the cross-configuration main set. (a) Distribution of frames across object classes. (b) Distribution of frames between daytime and nighttime conditions. (c) Distribution of valid annotated frames across LiDAR configurations. (d) Distribution of valid frames across distance intervals for the four LiDAR configurations.

TABLE IV
DISTRIBUTION OF DAYTIME AND NIGHTTIME DATA ACROSS EXPERIMENTAL VARIABLES IN THE CROSS-CONFIGURATION MAIN SET

| Variable | Level | Day | Night | Total | Night (%) |
|---|---|---|---|---|---|
| LiDAR | 16-beam | 699 | 446 | 1,145 | 39.0 |
| | 32-beam | 1,042 | 576 | 1,618 | 35.6 |
| | 80-beam | 2,577 | 1,387 | 3,964 | 35.0 |
| | 128-beam | 2,434 | 1,336 | 3,770 | 35.4 |
| Object class | Car | 2,785 | 2,669 | 5,454 | 48.9 |
| | SUV | 1,322 | 1,076 | 2,398 | 44.9 |
| | E-bike | 1,465 | 0 | 1,465 | 0.0 |
| | Bicycle | 1,180 | 0 | 1,180 | 0.0 |
| Distance | 0–20 m | 2,518 | 1,385 | 3,903 | 35.5 |
| | 20–40 m | 1,709 | 977 | 2,686 | 36.4 |
| | 40–60 m | 1,271 | 724 | 1,995 | 36.3 |
| | 60–80 m | 1,254 | 659 | 1,913 | 34.4 |
| Occlusion | Non-occluded | 5,481 | 2,404 | 7,885 | 30.5 |
| | Occluded | 1,271 | 1,341 | 2,612 | 51.3 |
| **Total** | | **6,752** | **3,745** | **10,497** | **35.7** |

Fig. 6(c) shows the distribution of valid annotated target frames across different LiDARs. Frame validity is determined using the uniform criterion defined in Section III-B, at least 40 LiDAR points within the refined target 3D box. The high-beam LiDARs (80- and 128-beam) contribute 3,964 and 3,770 valid frames, respectively. In contrast, the lower-beam LiDARs (16- and 32-beam) yield 1,145 and 1,618 valid frames, respectively. Because the same annotation workflow and valid-frame criterion were applied to all configurations, the lower valid-frame counts are consistent with the earlier loss of stably annotatable target geometry at longer distances for the lower-beam configurations.

Fig. 6(d) illustrates the distribution of valid frames across different distance intervals for the four LiDAR configurations. High-beam LiDARs (80- and 128-beam) maintain relatively consistent coverage across all distance ranges, whereas low-beam LiDARs (16- and 32-beam) exhibit a sharp decline in valid annotations beyond 20 m.

In addition to the cross-configuration main set, three deployment and operating condition subsets were collected under the same annotation and preprocessing procedure, each isolating one acquisition variable and analyzed separately in Section IV: a mounting-height subset (12 sequences, 1,256 frames, at 2.5 m mounting height; Section IV-C), a target-speed subset (8 sequences, 690 frames, at 30 and 50 km/h across all four LiDAR configurations; Section IV-D), and a curved-road subset (32 sequences, 7,288 frames, covering all four LiDAR configurations; Section IV-E), with the target-speed subset complementing the 10 km/h Car sequences already present in the cross-configuration main set. Each Section IV analysis draws on the portion of its subset relevant to the tested condition (e.g., the 128-beam Car sequences within the target-speed subset) rather than the full subset. These subsets are not included in the frame and sequence counts reported above or in the statistics of Fig. 6, which characterize the cross-configuration main set only.

# IV. LiDAR Diagnostics: Point Cloud Acquisition Performance

## A. *Distance-Dependent Geometric Degradation*

### 1) Comparison of LiDAR Configurations

Fig. 7(a) plots the target point count for the Car as a function of distance across the four LiDAR configurations in the cross-configuration main set. In general, the target point count decreases with distance, which is consistent with the theoretical LiDAR performance model [24]. The horizontal threshold in Fig. 7(a) corresponds to the valid-frame criterion defined in Section III-B: at least 40 LiDAR points on targets.

Under this criterion, valid annotated Car frames from the 16-beam LiDAR extend to approximately 20 m, whereas those from the 128-beam LiDAR extend beyond 80 m. The 16- and 32-beam LiDARs exhibit earlier geometric degradation with increasing distance, causing targets to degenerate into ambiguous point clusters with insufficient geometric detail, as shown in Fig. 7(c).

In contrast, the 80- and 128-beam LiDARs remain robust against increasing distance, maintaining usable geometric representations across the full measured range. Notably, the 80- and 128-beam configurations exhibit nearly identical target point counts across the measured distance range. This similarity is consistent with the measured beam profiles in Section III-C: both configurations provide 77 beams below the horizon, yielding nearly identical downward beam layouts and identical $k_{eff}$ values for Car across the evaluated distance intervals under the reference acquisition geometry of the cross-configuration main set. A shared feature distinguishing the two higher-beam configurations from the 16- and 32-beam configurations is their concentration of beams within the depression-angle band occupied by road targets. Nominal beam count alone is therefore insufficient to characterize target-relevant vertical scanning, and vertical beam distribution is also critical, consistent with the joint optimization findings in [19].

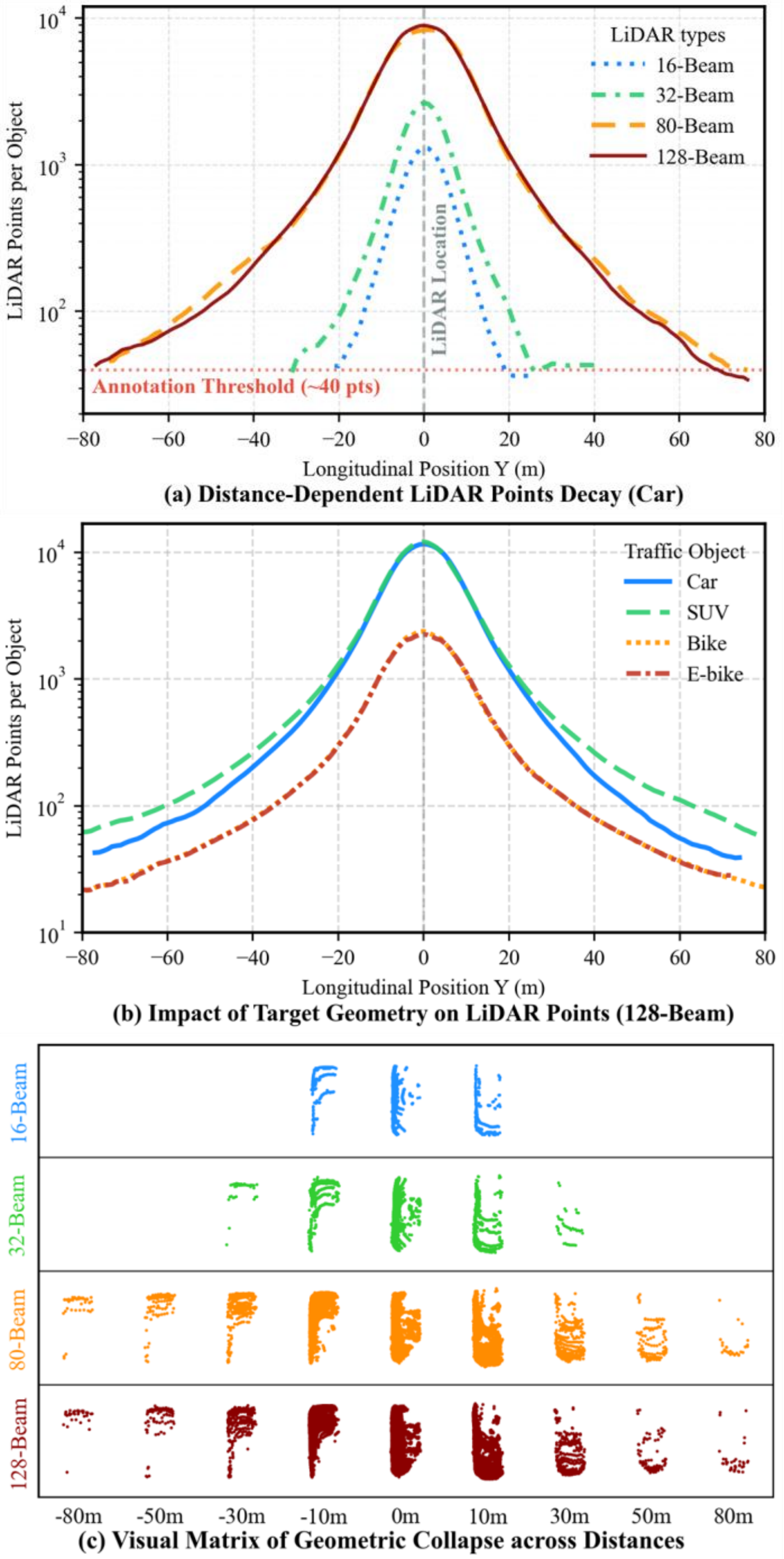

**Figure 7:** Quantitative Evaluation of LiDAR Perception Performance across Distances and Object Classes. (a) Point count on a Car target versus longitudinal position for the four LiDAR configurations. (b) Point count versus longitudinal position for the four object classes under the 128-beam LiDAR. (c) Point clouds of the same Car target at nine longitudinal positions from −80 m to 80 m under daytime, non-occluded conditions.

### 2) Comparison of Object Classes

Fig. 7(b) shows the number of LiDAR points on different object classes at varying distances for the 128-beam LiDAR.

For all object classes, point counts decrease with increasing distance. Taken together with the cross-configuration comparison above, this result indicates that geometric degradation depends jointly on LiDAR configuration and target geometry.

Within the motorized vehicle category, SUVs consistently yield more LiDAR points than passenger cars, especially at medium and long ranges. This advantage is attributed to the larger exposed surface area and higher vertical profile of SUVs.

A more pronounced discrepancy appears between motorized vehicles and VRUs. Compared with cars and SUVs, bicycles and e-bikes generate substantially fewer LiDAR points across the entire measured range. Owing to their slender geometry and limited exposed surface area, these two-wheeled targets suffer much more severe point cloud sparsity, making them inherently more difficult to represent reliably. Notably, the bike and e-bike curves are nearly overlapped, suggesting that their LiDAR scanning characteristics are highly similar. Quantitatively, the point count of a bicycle at approximately 20 m is comparable to that of a Car or SUV at approximately 40 m, highlighting the intrinsic challenge of perceiving small road users in roadside sensing.

**3) Comparison of Illumination Conditions**

To assess whether illumination condition affects target-point-count diagnostics, we conducted position-matched comparisons between daytime and nighttime observations for Car and SUV trajectories of the cross-configuration main set. The dense trajectory sampling obtained at the collection speed of 10 km/h enabled frame-level nearest-neighbor matching between daytime and nighttime observations. Within each LiDAR–object-class stratum, frames were matched using an *X*-position tolerance of 0.15 m, yielding eight non-occluded strata with 106–493 paired frames each.

Across all eight strata, the median daytime-to-nighttime point-count ratio fell within 0.98–1.04, indicating a typical deviation of less than 4%. As illustrated in Fig. 8, the daytime and nighttime point-count curves overlap closely across all four LiDAR configurations for both Car and SUV. These results indicate that, under the controlled conditions of this benchmark, daytime–nighttime illumination differences had little effect on target point counts.

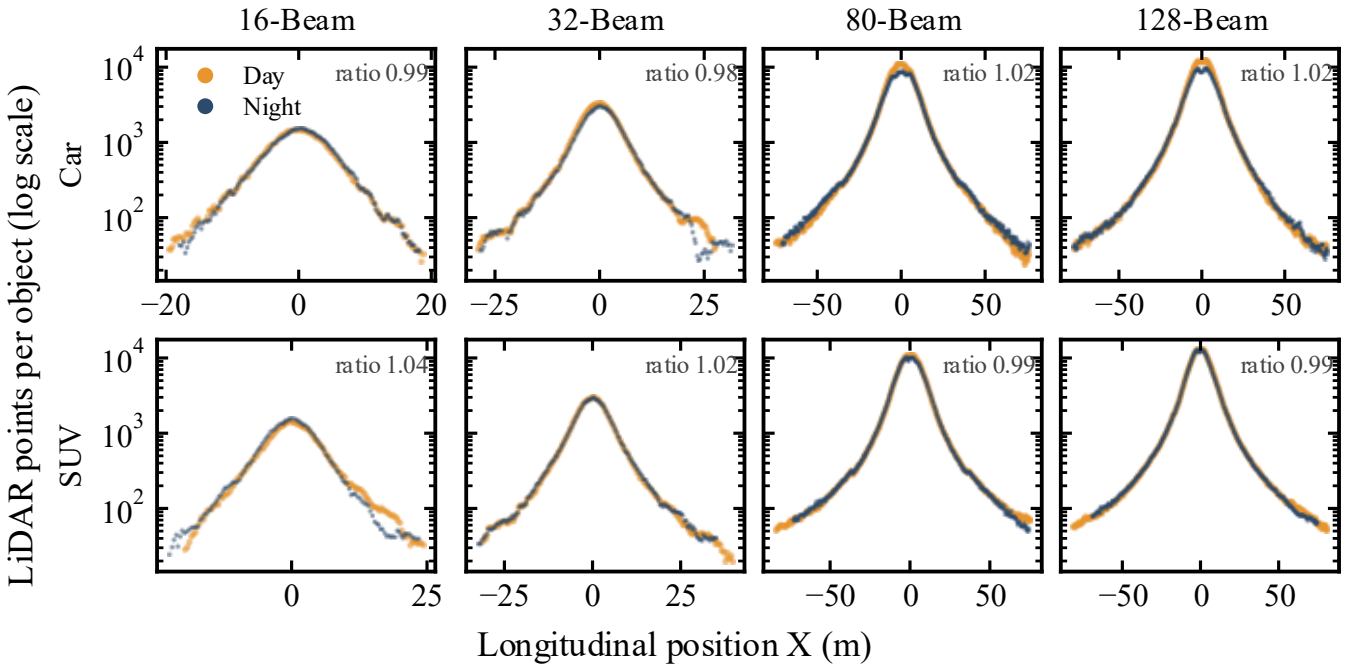

**Figure 8**: Position-matched comparison of daytime and nighttime point counts. Each panel shows one LiDAR configuration; daytime and nighttime frames are matched by longitudinal position *X* within a 0.15 m tolerance for non-occluded Car and SUV trajectories.

*B. Fine-Grained Occlusion Quantification*

Occlusion is another physical constraint that reduces the geometric information available in LiDAR point clouds. As illustrated in Fig. 9(a), the occluded and non-occluded scenarios in the cross-configuration main set were designed to differ only in the presence of a stationary obstacle vehicle between the target vehicle and the LiDAR. Existing LiDAR datasets generally do not provide a quantitative measure of occlusion-induced point loss. To address this limitation, MR-LiDAR introduces a point-based occlusion metric, termed the Occlusion Ratio ($O_{ratio}$). Specifically, $O_{ratio}$ is defined as the fraction of LiDAR points lost due to occlusion:

$$O_{ratio} = max\left(0,1-\frac{N_{occ}}{N_{ref}}\right)$$

where $N_{occ}$ denotes the measured point count under occlusion, and $N_{ref}$ denotes the point count obtained from the non-occluded reference sequence at the corresponding coordinate.

The non-occluded sequence serves as the baseline. To preserve the complete progression of occlusion-induced point loss, this sequence-level analysis uses the aligned occlusion trajectories before application of the benchmark-validity filtering described in Section III-B.

To obtain a consistent reference count, the occluded and non-occluded LiDAR sequences are aligned through curve registration. Specifically, the reference curve is rigidly shifted along the longitudinal coordinate *X* to minimize the residual with respect to the visible portion of the occluded sequence outside the occlusion zone:

$$\acute{\Delta} = arg\min_{\Delta\in[-3\text{m},3\text{m}]}\sum_{j\in V}\left(N_j^{occ} - g_{\Delta}\left(x_j^{occ}\right)\right)^2$$

where $g_{\Delta}(x)$ denotes the linear interpolation of the reference curve, shifted by $\Delta$, evaluated at position $x$, and $V$ denotes the visible frames of the occluded sequence outside the occlusion zone. The optimal shift $\acute{\Delta}$ is obtained by bounded scalar minimization.

As summarized in Table V, the sensitivity analysis was conducted across the aligned pre-filter frames in the partially occluded trajectories. The 95th percentile of the absolute deviation does not exceed 7.8 percentage points under a ±0.3 m alignment perturbation. Under 0.1 m random positional noise, the corresponding deviation does not exceed 3.5 percentage points based on 1,000 Monte Carlo trials per configuration. The 16-beam configuration exhibits the greatest sensitivity in both tests. These results indicate that the effect of the tested coordinate uncertainty is bounded but configuration-dependent, with the lower-beam LiDARs exhibiting greater sensitivity to positional perturbations.

Across all four LiDAR configurations, the target re-emerges from occlusion at a similar position, with the $O_{ratio}$ crossing occurring at 8.86 ± 0.55 m. This consistency suggests that the occlusion boundary is primarily determined by the fixed obstacle-vehicle geometry and is largely independent of beam count. It further supports the interpretation that $O_{ratio}$ reflects the physical occlusion process rather than sensor-specific noise.

Fig. 9(b) summarizes the distribution of frame-level $O_{ratio}$

across partially occluded frames, characterizing the severity of relative point loss within each LiDAR configuration. The median $O_{ratio}$ values are 0.52, 0.90, 0.80, and 0.78 for the 16-, 32-, 80-, and 128-beam LiDARs, respectively. Fig. 9(c) compares the point-count curves of the four LiDAR configurations. Point counts decrease substantially within the occlusion region. The 16- and 32-beam LiDARs yield only single-digit to low double-digit point counts during the most severely occluded portions, indicating near-complete loss of usable target geometry. In contrast, the 80- and 128-beam LiDARs intermittently retain tens to hundreds of points, particularly near the occlusion boundaries, although some deeply occluded frames also fall to single-digit counts. These results show that beam count affects the amount of residual geometric information available during partial occlusion.

TABLE V
SENSITIVITY OF $O_{ratio}$ TO ALIGNMENT AND POSITIONING UNCERTAINTY

| Beam | ±0.3 m (95th pct \|Δ\|) | Random 0.1 m (95th pct \|Δ\|) |
|---|---|---|
| 16 | 0.0783 | 0.035 |
| 32 | 0.0693 | 0.0279 |
| 80 | 0.0377 | 0.0131 |
| 128 | 0.0394 | 0.0129 |

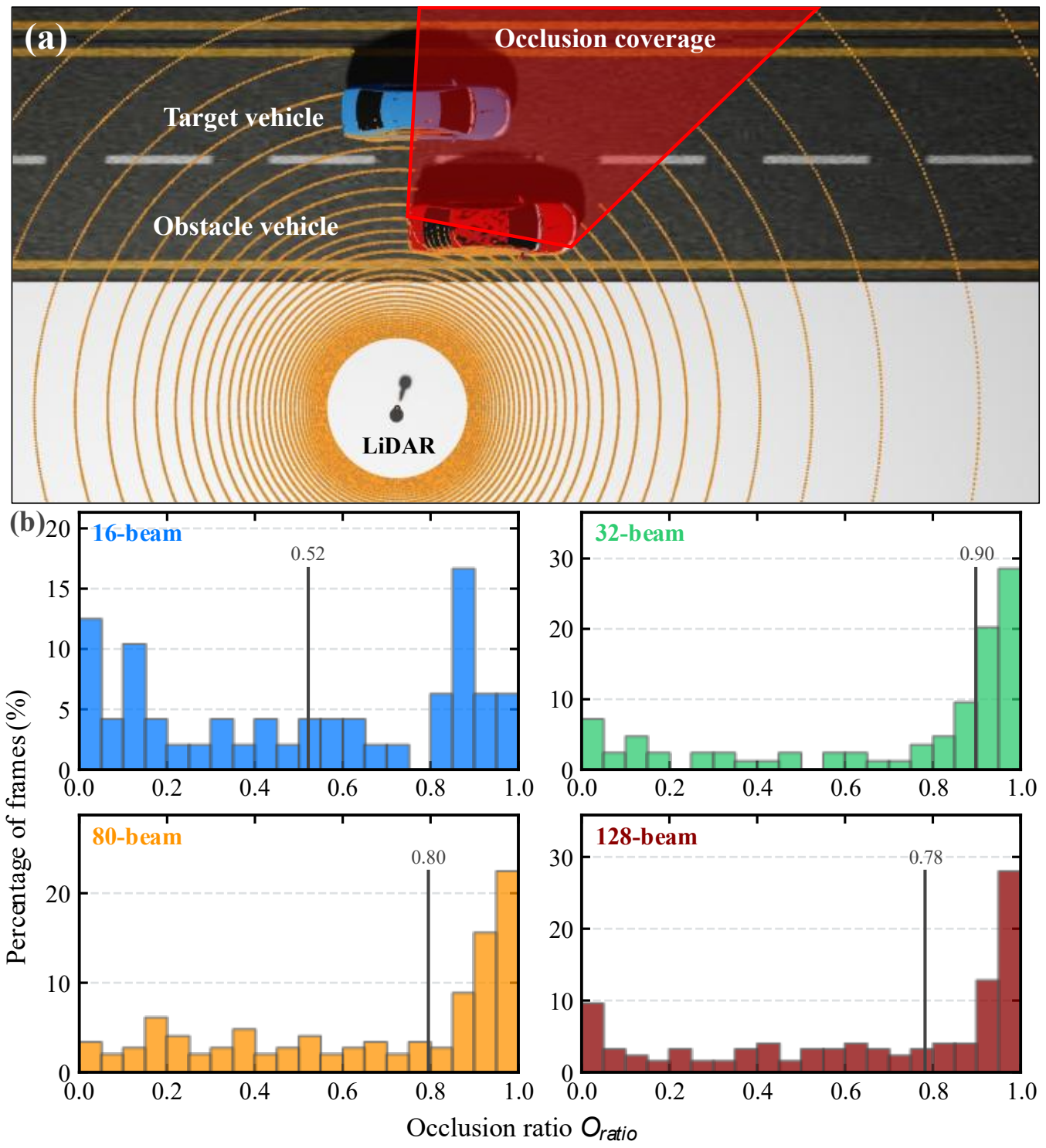

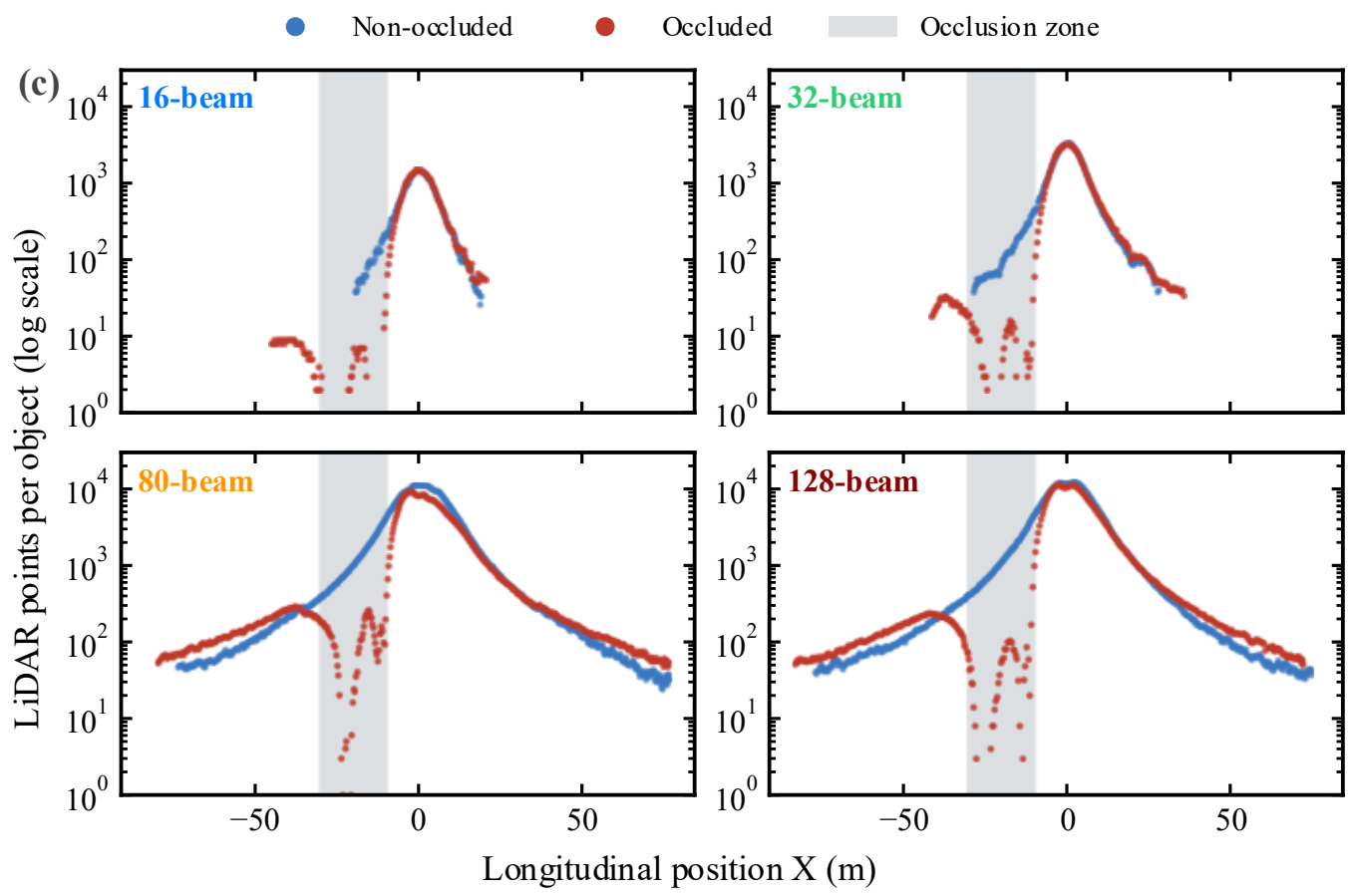

**Figure 9**: Fine-grained quantification of the point-level Occlusion Ratio. (a) Occluded and non-occluded acquisition scenarios, differing only in the presence of a stationary obstacle vehicle. (b) Distribution of frame-level $O_{ratio}$, showing the relative point-loss severity within each LiDAR configuration. (c) Point-count curves of the four LiDAR configurations across the occlusion region.

## C. Sensitivity to Mounting Height

Using the mounting-height subset introduced in Section III, we assess the sensitivity of the geometric diagnostics to the tested mounting-height change from 1.5 m to 2.5 m.

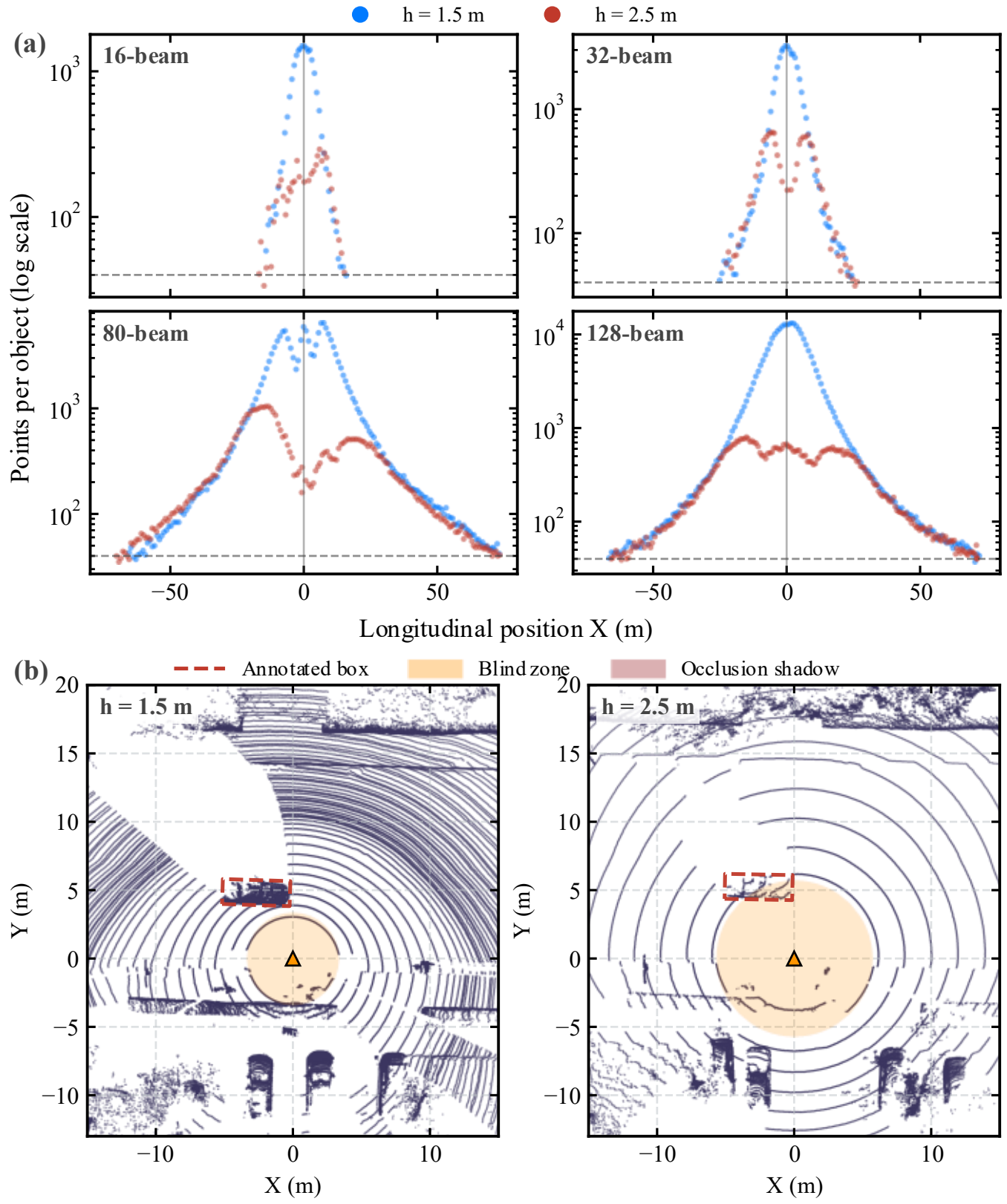

**Figure 10**: Effect of mounting height on point acquisition, for Car at a nominal speed of 30 km/h. (a) Frame-level point count versus longitudinal position X at 1.5 m and 2.5 m mounting heights, for the four LiDAR configurations. (b) Full-scene

bird's-eye-view point clouds of the 128-beam LiDAR at the closest range, comparing the enlarged near-field blind zone and the shortened occlusion shadow at 2.5 m.

Fig. 10(a) compares the frame-level point count as a function of longitudinal position $X$ for the two mounting heights, across all four LiDAR configurations. Two patterns are evident. First, in the far-field portion of each curve, the point-count curves obtained at 1.5 m and 2.5 m nearly coincide for all four LiDAR configurations. This indicates that the far-field point-count profiles are largely unaffected by the tested mounting-height change.

Second, in the near-field area, the two curves diverge substantially. The peak point count at 2.5 m is markedly lower than at 1.5 m. This near-field pattern is consistent with a fixed geometric constraint imposed by the vertical field of view. Because a LiDAR has a finite maximum depression angle, the ground beneath the LiDAR falls outside the coverage of any beam. Denoting the maximum depression angle as $\theta_{max}$ and the mounting height as h, the radius of this blind zone is given by

$$r_{blind} = \frac{h}{tan(|\theta_{max}|)}$$

This relation indicates that increasing the mounting height proportionally enlarges the near-field blind-zone radius. The maximum depression angle of the 128-beam LiDAR is −24.85°, giving $r_{blind}$ = 3.24 m at 1.5 m height and $r_{blind}$ = 5.40 m at 2.5 m height. Fig. 10(b) illustrates this effect through a bird's-eye-view comparison for the 128-beam LiDAR.

Fig. 10(b) also shows the effect of mounting height on the ground-return occlusion shadow. For the 128-beam example, this shadow is measured along the line of sight from the LiDAR through the target center and is defined as the distance immediately behind the vehicle over which no ground returns are observed. It extends approximately 10.0 m for 1.5 m LiDAR height but only 4.6 m for 2.5 m LiDAR height. This is consistent with the steeper depression angle at which the beams pass over the vehicle at greater mounting height, allowing the line of sight to be regained closer behind the target. Taken together, within the tested height range, raising the mounting height leaves the far-field point-count profiles largely unchanged but trades reduced near-field point count for a shorter ground-return occlusion shadow.

## D. Sensitivity to Target Speed

To examine whether target speed introduces speed-dependent bias into the geometric diagnostics, we analyze the target-speed subset, using the 128-beam Car sequences collected at 10, 30, and 50 km/h under the same daytime condition, mounting height, road segment, annotation procedure, and preprocessing procedure. As described in Section III-A, the 128-beam configuration was selected because its high angular resolution and target-point density provide a sensitive case for detecting speed-related effects, including scan-induced geometric distortion and reduced trajectory sampling density.

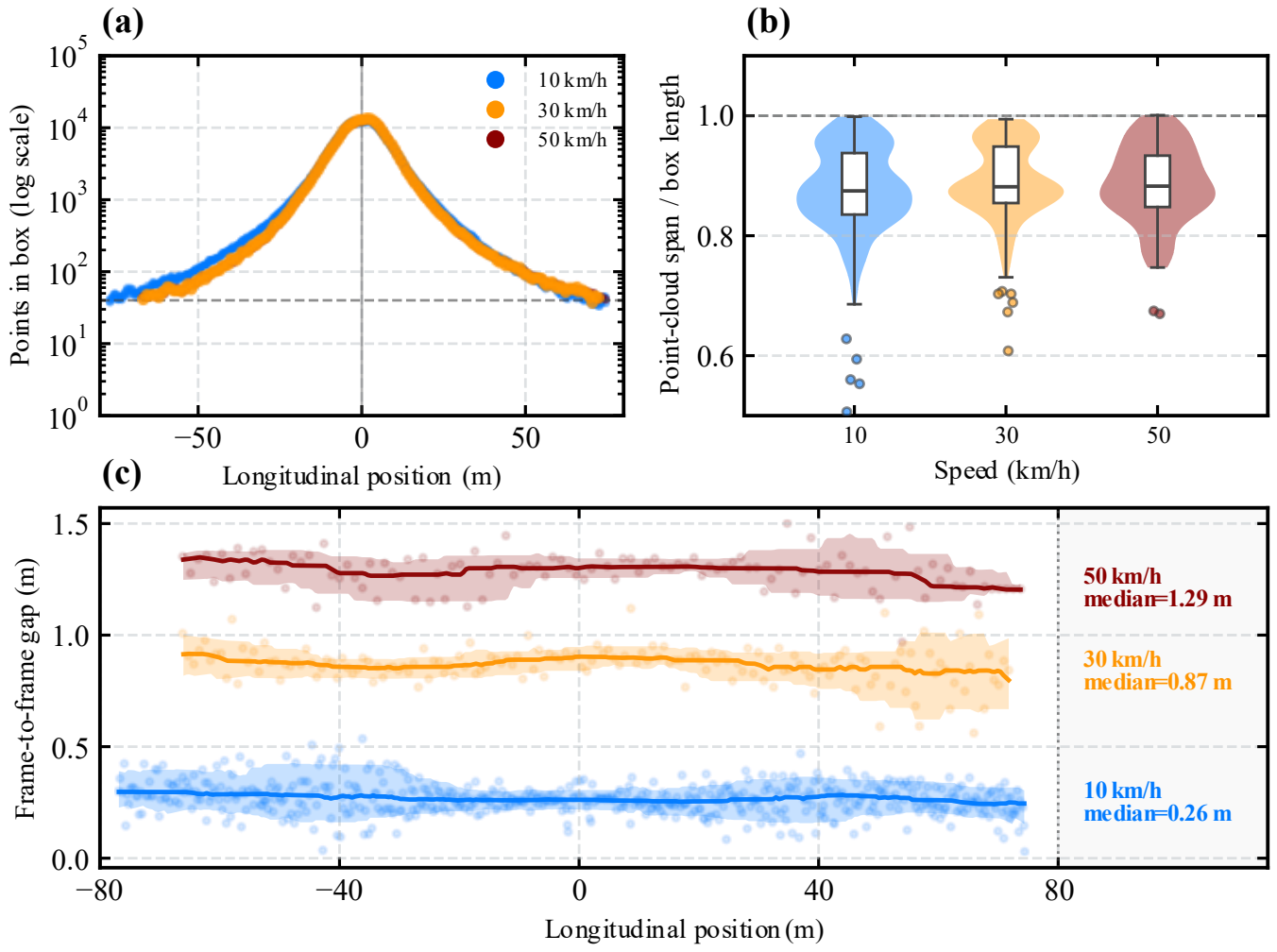

**Figure 11**: Sensitivity to target speed for the 128-beam LiDAR with a Car target at 1.5 m mounting height, comparing 10, 30, and 50 km/h. (a) Point count versus longitudinal position. (b) Motion-blur proxy, defined as the ratio of point-cloud longitudinal span to annotated box length. (c) Frame-to-frame sampling gap versus position.

Fig. 11(a) compares the target point count as a function of longitudinal position across the three speeds. The three curves closely overlap in peak magnitude and overall profile, indicating no evident speed-dependent change in the target-point-count patterns reported in Section IV-A within the tested range.

To assess motion blur, we compare the longitudinal span of the target point cloud with the annotated box length. A ratio near unity indicates that the point cloud spans approximately the same extent as the box, whereas a ratio substantially greater than one would indicate tailing caused by target displacement during the scanning cycle. As shown in Fig. 11(b), the median ratio is 0.875, 0.882, and 0.883 at 10, 30, and 50 km/h, respectively, with overlapping interquartile ranges across all three speeds. A small number of frames with span ratios below 0.7 are present at every speed level, and their proportion does not increase with speed. These results show no detectable speed-dependent increase in longitudinal point-cloud elongation within the tested range.

Sparse scanning was evaluated through the frame-to-frame gap, defined as the longitudinal distance between the centers of consecutive annotated boxes. Under a fixed frame rate, this gap is expected to scale linearly with target speed. Fig. 11(c) shows the median gap increasing from 0.26 m at 10 km/h to 0.87 m at 30 km/h and 1.29 m at 50 km/h, corresponding to ratios of 3.3× and 4.9× relative to the 10 km/h baseline. These ratios closely match the nominal speed ratios of 3× and 5×, confirming that the reduced scanning density at higher speeds is a direct and predictable consequence of the fixed acquisition frame rate.

Taken together, these results indicate that increasing target speed from 10 to 50 km/h does not alter the underlying distance-decay geometry and does not introduce detectable motion blur, while proportionally reducing the density of trajectory samples in a manner consistent with the fixed acquisition frame rate.

### *E. Sensitivity to Curved-Road Geometry*

To examine whether the findings obtained on the straight-road segment persist under road curvature, we analyze the curved-road subset collected on a curved segment under daytime conditions at a target speed of 10 km/h. The subset covers all four LiDAR configurations, Car and Bicycle targets and mounting heights of 1.5 m and 2.5 m. For each LiDAR–class–height combination, one left-turn and one right-turn traversal were recorded, yielding 32 sequences in total.

The curved segment is an L-shaped right-angle turn, with the LiDAR positioned near the inside of the corner. Here, only the turning arc is retained, defined by the azimuth window $\theta \in [45°, 225°]$ in the LiDAR *xy*-plane. The retained left-turn sequences contain 3,482 annotated frames within this azimuth window. In this setting, a turning target sweeps approximately 180° of azimuth relative to the sensor over a comparatively constrained range. This geometry allows the variation in target point count with viewing azimuth to be examined within a single traversal.

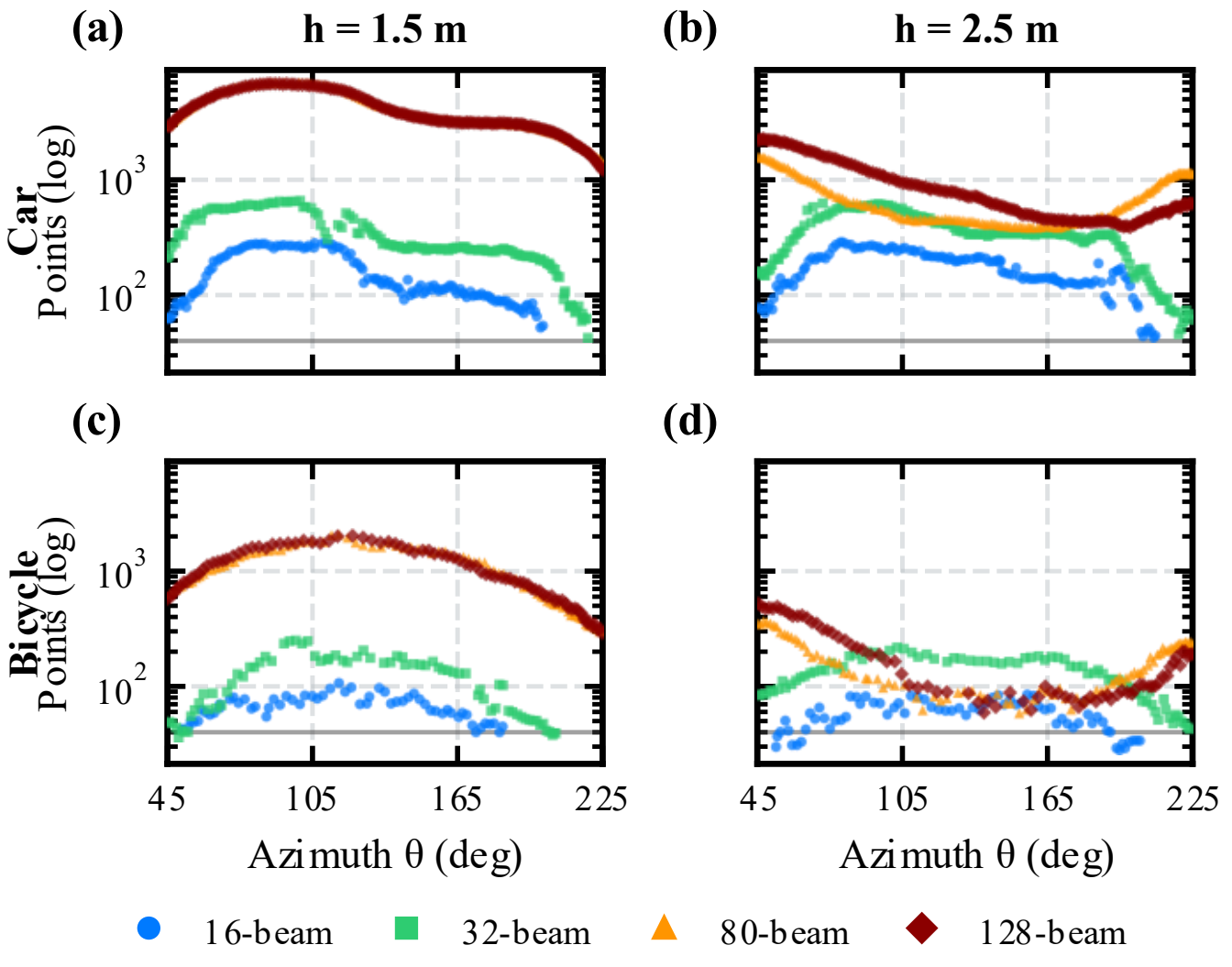

**Figure 12:** Per-frame target point count across the turning arc, plotted against target azimuth $\theta$ in the LiDAR *xy*-plane. Car and Bicycle targets at mounting heights of 1.5 m and 2.5 m.

Fig. 12 shows the per-frame target point count as a function of azimuth for the left-turn sequences, with two target classes at the two mounting heights. At a mounting height of 1.5 m, the four configurations separate into three distinct tiers. The 80-beam and 128-beam point clouds overlap closely across the entire sequence, with median counts of 3,380 and 3,248 points for Car, followed by the 32-beam at 254 points and the 16-beam at 116 points. This ordering reproduces the cross-configuration point-count pattern reported in Section IV-A for the straight-road segment, indicating that the ordering persists in the tested curved-road setting at a mounting height of 1.5 m.

At a mounting height of 2.5 m, the tier structure contracts substantially. For Car, the median count decreases from 3,380 to 537 points for the 80-beam LiDAR and from 3,248 to 611 points for the 128-beam LiDAR, while the 32-beam median increases from 254 to 332 points. For Bicycle, the 32-beam median of 130 points slightly exceeds the 80-beam median of 129 points, so the two tiers become indistinguishable. Compared with the substantial reductions observed for the 80- and 128-beam configurations, the point-count patterns of the 16- and 32-beam configurations change less markedly between the two heights. The height sensitivity is therefore considerably stronger for the two high-beam configurations in this curved-road subset.

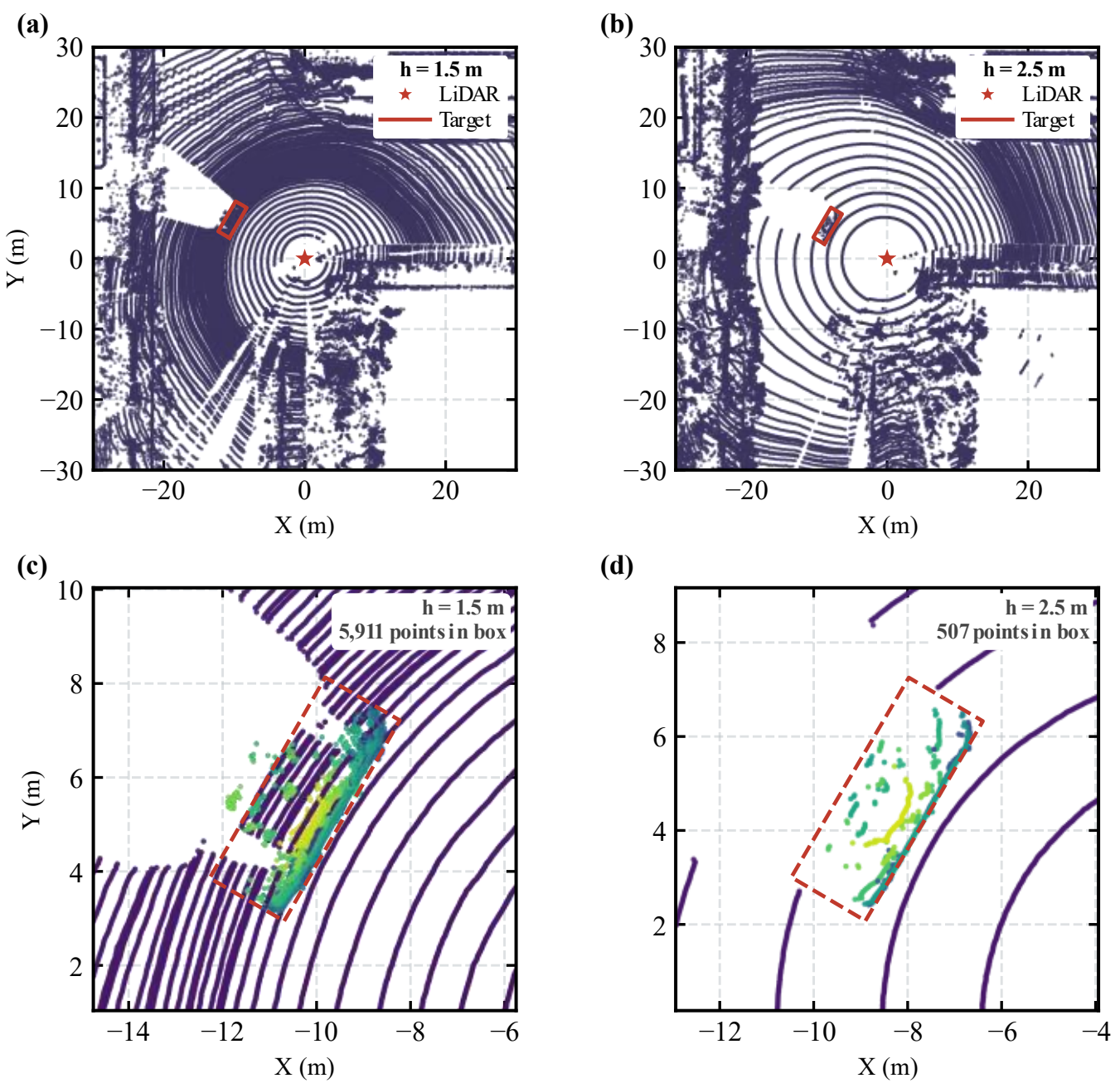

**Figure 13:** Curved-road point cloud for the 128-beam LiDAR on a right turn, matched at a horizontal range of 8.0 m. Full-scene views show the outward expansion of the concentric ground-return rings at the greater mounting height; close-up views show the corresponding reduction in points on the target.

The full-scene views (Fig. 13) show that the concentric ground-return rings surrounding the LiDAR expand outward at the greater mounting height, consistent with the blind-zone relation introduced in Section IV-C. The close-up views show the consequence for the target. At 1.5 m the vehicle body is covered by a dense and continuous point cloud containing 5,911 points, whereas at 2.5 m the same vehicle at the same range retains 507 points distributed over a small number of sparse scan lines. This single-frame comparison illustrates the mechanism.

The asymmetry between the two tiers follows from their vertical beam distributions, measured per beam in Section III-C. As reported in Table III, the 16-beam and 32-beam LiDARs have a uniform distribution. The 80-beam and 128-beam LiDARs have non-uniform distribution, with a median spacing of 0.100° and 50 beams concentrated in the narrow band between −5° and 0°.

A non-uniform distribution concentrates scanning capacity within a specific range of depression angles, so its benefit is realized only when the target subtends that range. Raising the LiDAR height increases the magnitude of the depression angle toward a nearby target and shifts the target's angular extent away from the densely sampled band. Uniform beam distributions are also affected by mounting height, but they do not exhibit the same concentrated loss associated with leaving a narrow high-density angular band. Applying the effective beam count $k_{eff}$ defined in Section III-C, for Car at a horizontal range of 7.5 m $k_{eff}$ is 6, 12, 72, and 72 for the four configurations at 1.5 m, and 4, 9, 7, and 7 at 2.5 m. At this near-field location, the two high-beam configurations therefore lose most of their effective-beam advantage. Although $k_{eff}$ does not fully determine the frame-level point count, it explains the disproportionate reduction in target-relevant vertical sampling for the two non-uniform configurations. These results refine the conclusion of Section IV-A. The advantage of an optimized non-uniform beam distribution depends on whether the target subtends the angular band in which the beams are concentrated. Under the tested elevated near-field condition, the point-count advantage of the 80- and 128-beam configurations narrows substantially; for Bicycle, the uniformly distributed 32-beam configuration provides a median point count comparable to that of the 80-beam configuration. However, the analysis is limited to the near field, as the retained turning arc lies within approximately 25 m of the LiDAR.

## V. LiDAR Diagnostics: Algorithmic Perception Performance

Complementing the geometric diagnostics in Section IV, this section evaluates how the observed cross-configuration differences translate into target-specific algorithmic detection performance.

### *A. Experimental Setup*

**1) Baselines:** To reduce dependence on a particular detector architecture and examine whether the cross-configuration patterns remain consistent across detection paradigms, we adopt two widely used 3D detectors, the anchor-free CenterPoint [25] and the classic anchor-based SECOND [26]. The two models share an identical voxel encoder and 3D backbone, comprising MeanVFE, VoxelBackBone8x, and a 256-dimensional height compression module, and differ only in the BEV backbone and detection head. SECOND employs a two-level BaseBEVBackbone with five blocks per level and 128 and 256 filters, followed by AnchorHeadSingle, whereas CenterPoint employs a single-level backbone with five blocks and 128 filters, followed by CenterHead with the four classes assigned to two groups, [Car, SUV] and [E-bike, Bicycle]. For a given detector, the point cloud range $[x_{min}, y_{min}, z_{min}, x_{max}, y_{max}, z_{max}]$ = [−80, 1.5, −1.5, 80, 7.9, 6.5] m, the voxel size of [0.05, 0.05, 0.2] m, and the class definitions are held constant across the four LiDAR configurations, allowing cross-LiDAR performance differences to be evaluated under a common input representation.

**2) Training:** We implemented a sensor-specific training strategy, whereby separate models were trained for each LiDAR configuration using its corresponding data from the cross-configuration main set. All point clouds were converted to KITTI format and divided into training, validation, and test sets in a 70%/15%/15% ratio. Both detection models were implemented within the OpenPCDet framework and trained for 30 epochs. A fixed 30-epoch training schedule was used for every detector–LiDAR combination and was determined before evaluation on the test set. We used the Adam OneCycle optimizer with a maximum learning rate of 0.003, a weight decay of 0.01, a momentum range of [0.95, 0.85], and a batch size of 12 per GPU. Standard data augmentations were applied, including random flipping, global rotation, scaling, and ground-truth sampling to balance class distribution. Every detector and LiDAR combination was trained under this identical schedule and evaluated at the final epoch-30 checkpoint, without checkpoint selection based on test-set results. The experiments were conducted with OpenPCDet 0.6.0 and PyTorch 2.11 on a single NVIDIA RTX 5070 GPU.

**3) Evaluation:** Because the benchmark consists of controlled, target-centric single-target sequences rather than fully annotated multi-object traffic scenes, standard scene-level mAP does not directly reflect the diagnostic objective of this study. Instead, we adopt a target-specific strategy using three complementary indicators.

- **Recall** measures the capability of discovering the target. We employ distance-based recall to evaluate target discovery, consistent with the target-association criterion used in trajectory reconstruction tasks [27]. It is defined as the percentage of annotated targets that are successfully detected:

$$Recall = \frac{N_{TP}}{N} \times 100\%$$

where $N_{TP}$ denotes the number of detected targets. A detection is a True Positive if the Euclidean distance between the predicted center $c_{pred}$ and ground truth center $c$ satisfies $\|c_{pred} - c\|_2 < \tau_{dist}$ with $\tau_{dist} = 2.0m$, $N$ denotes the number of annotated targets.

- **Average IoU** measures the geometric fidelity of the predicted bounding boxes:

$$AvgIoU = \frac{1}{N_{TP}} \sum_{i=1}^{N_{TP}} IoU_i$$

- **Average Translation Error (ATE)** evaluates localization precision. It is computed as the mean Euclidean distance between predicted and ground-truth centers over all matched true positives:

$$ATE = \frac{1}{N_{TP}} \sum_{k=1}^{N_{TP}} \|c_{pred}^k - c^k\|_2$$

The same process, metrics, and class definitions were applied to all four LiDAR configurations. The results in Section V-B therefore represent sensor-specific detection performance under common model configurations, input representations, training schedules, and evaluation rules. They are used to assess whether cross-configuration trends remain consistent across the two detectors.

### *B. Detection Performance Across Configurations, Classes, and Ranges*

Table VI reports the results of the two detectors using three metrics across LiDAR configurations, object classes, and distance intervals. These results provide an empirical basis for the configuration guidance developed in Section V-C.

**1) Perception Performance of Different LiDARs:** Across the four configurations, denser target-relevant sampling is generally associated with stronger detection and localization performance. In particular, the 16-beam LiDAR exhibits degraded localization accuracy even at very close ranges, with an ATE of 0.226m/0.310 m. More importantly, its performance deteriorates sharply beyond 20 m: the Recall for cars drops to 17.27%/4.55% in the 20–40 m range, while the localization error increases to 0.767 m/0.724 m.

TABLE VI
QUANTITATIVE LIDAR DIAGNOSTICS BENCHMARK RESULTS

| CenterPoint / SECOND | Range | 0-20m | | | 20-40m | | | 40-60m | | | 60-80m | | |
|---|---|---|---|---|---|---|---|---|---|---|---|---|---|
| **LiDAR Model** | **Object Class** | **Recall** | **IoU** | **ATE** | **Recall** | **IoU** | **ATE** | **Recall** | **IoU** | **ATE** | **Recall** | **IoU** | **ATE** |
| **128-Beam** | **Car** | 90.32/90.11 | 84.66/86.72 | 0.114/0.099 | 89.54/91.21 | 85.28/85.47 | 0.113/0.122 | 99.80/100.00 | 83.24/81.10 | 0.118/0.148 | 98.91/96.51 | 77.09/77.07 | 0.151/0.17 |
| | **SUV** | 100.00/100.0 | 87.39/91.33 | 0.069/0.062 | 100.00/100.00 | 88.66/89.70 | 0.073/0.076 | 100.00/100.00 | 88.03/85.93 | 0.071/0.108 | 99.53/99.53 | 82.57/83.74 | 0.113/0.118 |
| | **E-bike** | 100.00/100.00 | 83.52/84.99 | 0.056/0.070 | 100.00/99.26 | 83.25/78.63 | 0.055/0.124 | 100.00/100.00 | 81.47/79.18 | 0.076/0.101 | 86.44/83.90 | 73.60/72.69 | 0.137/0.147 |
| | **Bicycle** | 100.00/100.00 | 73.01/80.71 | 0.077/0.105 | 100.00/100.00 | 79.02/82.37 | 0.088/0.088 | 100.00/100.00 | 80.55/78.93 | 0.091/0.106 | 100.00/100.00 | 79.90/76.72 | 0.103/0.122 |
| **80-Beam** | **Car** | 88.25/92.87 | 86.12/84.44 | 0.098/0.166 | 89.76/94.88 | 81.18/82.38 | 0.180/0.222 | 93.13/98.28 | 76.78/78.61 | 0.240/0.251 | 89.54/89.34 | 71.25/74.35 | 0.251/0.222 |
| | **SUV** | 100.00/100.00 | 85.98/90.51 | 0.067/0.068 | 99.53/100.00 | 86.42/88.89 | 0.087/0.084 | 100.00/100.00 | 84.88/87.05 | 0.094/0.095 | 100.00/100.00 | 79.67/82.14 | 0.131/0.129 |
| | **E-bike** | 98.06/100.00 | 84.94/78.65 | 0.055/0.099 | 98.03/100.00 | 83.97/80.66 | 0.060/0.092 | 100.00/100.00 | 81.33/79.73 | 0.066/0.098 | 100.00/100.00 | 76.27/78.26 | 0.120/0.110 |
| | **Bicycle** | 94.96/100.00 | 82.88/70.83 | 0.063/0.165 | 98.33/100.00 | 78.20/74.71 | 0.102/0.125 | 100.00/100.00 | 77.08/75.66 | 0.092/0.119 | 100.00/98.84 | 76.14/76.64 | 0.115/0.121 |
| **32-Beam** | **Car** | 87.35/88.78 | 83.33/80.02 | 0.120/0.164 | 79.08/82.77 | 69.45/69.06 | 0.293/0.279 | 66.67/83.33 | 40.81/43.3 | 0.983/0.838 | - | - | - |
| | **SUV** | 100.00/100.00 | 87.21/72.81 | 0.076/0.256 | 100.00/98.16 | 82.06/80.79 | 0.124/0.152 | - | - | - | - | - | - |
| | **E-bike** | 99.31/100.00 | 78.27/74.29 | 0.085/0.123 | 100.00/100.00 | 73.55/72.97 | 0.133/0.150 | - | - | - | - | - | - |
| | **Bicycle** | 99.21/96.06 | 73.74/67.24 | 0.114/0.145 | 100.00/100.00 | 72.58/71.03 | 0.153/0.161 | - | - | - | - | - | - |
| **16-Beam** | **Car** | 89.39/84.73 | 71.02/67.81 | 0.226/0.310 | 17.27/4.55 | 42.40/38.18 | 0.767/0.724 | | | | - | - | - |
| | **SUV** | 99.58/99.58 | 70.95/73.92 | 0.169/0.201 | 86.67/96.67 | 54.69/64.46 | 0.418/0.341 | - | - | - | - | - | - |
| | **E-bike** | 99.12/95.58 | 69.69/55.49 | 0.096/0.313 | - | - | - | - | - | - | - | - | - |
| | **Bicycle** | 99.02/82.35 | 61.46/24.49 | 0.165/0.757 | - | - | - | - | - | - | - | - | - |

Note: To demonstrate cross-algorithm generalizability, quantitative metrics in Table VI are presented in the format of CenterPoint / SECOND.

The simultaneous reduction in Recall and increase in ATE indicate a marked loss of target discovery and localization reliability beyond 20 m under the tested setting. The 32-beam LiDAR also exhibits clear performance degradation at mid-range distances; for example, the Car IoU decreases to 69.45%/69.06% in the 20–40 m interval. By contrast, the 80- and 128-beam configurations generally retain higher Recall and localization quality across the evaluated distance intervals. These results show that LiDAR configuration, including nominal beam count and target-relevant vertical beam distribution, is strongly associated with target-specific detection performance.

**2) Perception Performance for Different Object Classes:** The results reveal clear object-class-dependent differences in LiDAR perception performance. Overall, SUVs generally achieve the highest Recall and Average IoU across the evaluated configurations and distance intervals, a pattern consistent with their larger dimensions and more extensive visible surfaces. Cars exhibit slightly weaker robustness, especially under low-resolution LiDARs at mid-to-long distances, where point sparsity leads to noticeable degradation in both overlap quality and localization accuracy. For instance, under the 32-beam LiDAR, the Car IoU drops from 83.33%/80.02% at 0–20 m to 69.45%/69.06% at 20–40 m, and further to 40.81%/43.3% at 40–60 m, with the ATE increasing sharply to 0.983 m/0.838 m.

By contrast, VRUs, including e-bikes and bicycles, are generally more challenging to perceive than motor vehicles, particularly in terms of geometric completeness and localization stability. Although high-resolution LiDARs can still maintain high Recall for these classes, their IoU is often lower than that of SUVs and sometimes lower than that of cars, indicating that small and slender objects are more vulnerable to point cloud sparsity. For example, with the 128-beam LiDAR at 0–20 m, bicycles achieve an IoU of only 73.01%/80.71%, compared with 87.39%/91.33% for SUVs. These findings indicate that target geometry is an additional source of variation in detection performance: larger motor vehicles generally yield stronger detection and localization results than smaller and more slender traffic participants.

**3) Association with Effective Beam Count:** The results further show that nominal beam count alone does not explain the cross-configuration detection patterns. Although the 80-beam configuration has 37.5% fewer nominal beams than the 128-beam configuration, the two achieve broadly comparable results across object classes and distance intervals under the reference acquisition geometry of the cross-configuration main set. This pattern is consistent with their nearly identical downward beam layouts characterized in Section III-C.

To examine whether this geometric account extends to detector output, the effective beam count of Section III-C was matched to the results of Table VI. For each valid LiDAR-configuration–object-class–distance cell in Table VI, the median $k_{eff}$ was paired with distance-based Recall, Average IoU, and ATE. Separate weighted cell-level regressions were fitted for the two detectors, with object class and distance interval included as control variables and the number of ground-truth targets used as the cell weight. This yielded 47 cells with valid values for all variables. Each additional effective beam is associated with an increase in Average IoU of 0.177 percentage points for CenterPoint ($p = 1.2 \times 10^{-4}$) and 0.280 percentage points for SECOND ($p = 2.6 \times 10^{-6}$), and with a reduction in ATE of 0.0015 m ($p = 0.0097$) and 0.0026 m ($p = 2.1 \times 10^{-4}$). Across the observed $k_{eff}$ range for Car at 0–20 m, from 4.5 for the 16-beam configuration to 67.5 for the 80-beam configuration, these coefficients correspond to model-implied Average IoU differences of 11.2 percentage points for CenterPoint and 17.6 percentage points for SECOND. The association with Distance Recall is weaker and reaches significance for neither detector ($p = 0.364$ and $p = 0.107$), consistent with Recall approaching saturation for the high-beam configurations at most ranges. Within the reference acquisition geometry of the cross-configuration main set, higher $k_{eff}$ is therefore more consistently associated with matched-box overlap and localization accuracy than with target discovery.

## *C. Practical LiDAR Selection Guide*

Building on the empirical findings in Sections IV and V, we construct a minimal-sufficient LiDAR configuration matrix that maps object class and target distance to the lowest-beam tested configuration satisfying two operational criteria (Fig. 14).

**1) Dual-criterion definition:** A configuration is recommended for a given (object class, distance interval) only if it satisfies two complementary conditions: (i) frame-level geometric sufficiency, i.e., at least 70% of all candidate frames acquired within the interval before valid-frame filtering retain ≥40 points on the target, where the denominator comprises all candidate frames in that interval; (ii) matched-box quality, i.e., the CenterPoint Average IoU for that cell in Table VI reaches ≥70%. The point-count criterion measures whether sufficient target geometry is retained across the candidate frames, whereas Average IoU measures the box overlap achieved for successfully matched detections. Requiring both criteria prevents selection based solely on either geometric availability or matched-box quality. Among the four tested configurations satisfying both conditions for a given cell, the configuration with the lowest nominal beam count is reported as minimal-sufficient. A cell is marked N/A when none of the tested configurations satisfies both criteria.

**2) Resulting matrix:** Across the 16 (object class × distance) cells, 14 admit a minimal-sufficient recommendation and 2 are marked N/A. The selected nominal beam count is generally nondecreasing with distance within each class, consistent with the distance-dependent geometric degradation reported in Section IV-A. For motorized vehicles, the 16-beam LiDAR is sufficient at 0–20 m, while Car requires the full 128-beam configuration beyond 60 m; SUV, benefiting from its larger exposed surface, remains adequately served by the 80-beam LiDAR up to 80 m. For VRUs, the matrix selects at least the 32-beam configuration at 0–20 m and the 80-beam configuration at 20–60 m. At 60–80 m, no tested configuration satisfies both criteria for either E-bike or Bicycle. The non-uniform 80-beam configuration is selected in several mid-to-far-range cells, showing that the additional upward beams of the 128-beam configuration are not required to satisfy the two adopted criteria in those cells under the reference geometry of the cross-configuration main set.

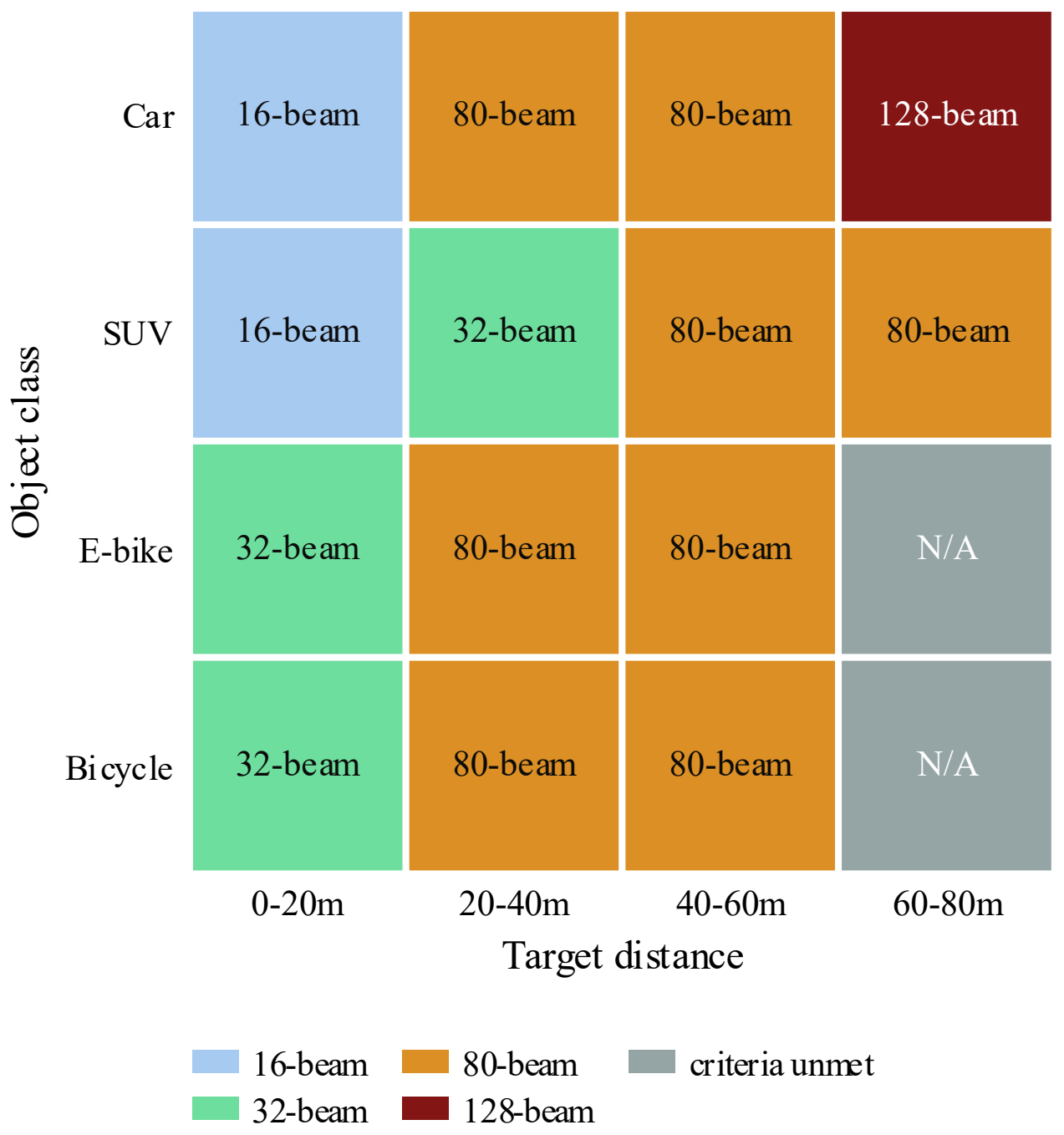

**Figure 14:** Minimal-sufficient LiDAR configuration matrix by object class and target distance. Each cell reports the lowest-beam tested configuration satisfying both criteria: at least 70% of all candidate frames in the distance interval contain ≥40 target points before valid-frame filtering, and CenterPoint Average IoU is ≥70%.

**3) Threshold sensitivity and scope:** The frame-sufficiency and Average IoU thresholds of 70% are empirical operational choices. A grid sensitivity analysis was conducted by varying both thresholds from 60% to 90%. Within the region defined by frame-sufficiency thresholds of 60–85% and Average IoU thresholds of 60–75%, each resulting matrix differs from the reported matrix in no more than three of the 16 cells (0–19%). Varying the frame-sufficiency threshold from 60% to 85% does not change the number of cells for which at least one tested configuration is admissible, although it may change which configuration is selected as minimal-sufficient. The matrix becomes substantially more restrictive when the Average IoU threshold exceeds 80%, with 56–88% of the cells changing relative to the reported matrix. The two adopted values should therefore be interpreted as empirical operational thresholds located within a locally stable region, rather than as universal detection criteria. The matrix is derived from the cross-configuration main set collected on a single 200 m straight-road segment at a mounting height of 1.5 m and a target speed of 10 km/h, using CenterPoint results as the algorithmic criterion. It should therefore be interpreted as an operational summary for the tested conditions of the cross-configuration main set rather than as a detector-independent or universal hardware recommendation. Sections IV-C and IV-E show that mounting height and near-field curved-road geometry can alter cross-configuration point-count patterns, and the matrix has not been recalibrated for those conditions. Although Section IV-D shows stable point-count profiles from 10 to 50 km/h for the tested 128-beam Car sequences, the matrix has not been validated using detection results at the higher speeds.

## VI. Conclusion

This paper presented MR-LiDAR, a controlled cross-configuration benchmark for roadside LiDAR perception diagnostics. Using matched observations across four LiDAR configurations and multiple object classes, it characterizes how target point availability and detection performance vary with sensor configuration, target geometry, distance, and acquisition conditions.

Three main findings are obtained. First, the tested 16- and 32-beam configurations exhibit earlier distance-dependent degradation in target point count, Recall, Average IoU, and ATE than the 80- and 128-beam configurations. Second, SUVs generally yield stronger geometric and detection results, whereas VRUs which are smaller and slender are more affected by point sparsity. Third, nominal beam count alone is insufficient to describe target-relevant vertical scanning. The 80- and 128-beam configurations have nearly identical downward beam layouts and achieve broadly comparable results under the reference acquisition geometry of the cross-configuration main set. Moreover, at the same nominal beam count and vertical field of view, the uniform 80-beam baseline provides approximately one third of the effective beam count of the actual non-uniform configuration. These findings support the effective beam count $k_{eff}$ as a more direct descriptor of target-relevant vertical scanning.

The proposed Occlusion Ratio quantifies sequence-level point loss during partial occlusion. The results show that the occlusion boundary is mainly determined by scene geometry, whereas the amount of target geometric information retained during occlusion depends on the LiDAR configuration. Moreover, daytime–nighttime illumination and the target speed have little influence on target point count patterns. Increasing mounting height reduces near-field sampling, and shortens the ground-return occlusion shadow, but enlarges the near-field blind zone. On the curved road, the configuration ranking at 1.5 m is consistent with that observed on the straight road, whereas the advantage of the high-beam configurations decreases substantially in the near field at 2.5 m.

Finally, the constructed minimal-sufficient configuration matrix identifies the lowest-beam tested configuration for each object-class and distance cell. It is an operational summary based on MR-LiDAR. Future work will extend MR-LiDAR to additional sites, sensor tilt angles, adverse weather, intersections, and dense multi-object traffic, and will further evaluate detection and tracking performance under these conditions.

## References

[1] C. Creß, Z. Bing, and A. C. Knoll, “Intelligent Transportation Systems Using Roadside Infrastructure: A Literature Survey,” *IEEE Trans. Intell. Transp. Syst.*, vol. 25, no. 7, pp. 6309–6327, Jul. 2024, doi: 10.1109/TITS.2023.3343434.

[2] S. Xu, C. Wang, F. Xing, and S. Ai, “Surface Displacement Measurement Method in Enclosed Underground Tunnel Based on Mobile Mapping,” *IEEE Trans. Instrum. Meas.*, vol. 75, p. 8510619, 2026, doi: 10.1109/TIM.2026.3709359.

[3] C. Wang, S. Xu, W. Li, Y. Wang, A. Song, and F. Xing, “Novel method for regional surface displacement measurement of coal mine roadway under different typical working conditions based on 3D laser scanning technique,” *Measurement*, vol. 261, p. 119870, Feb. 2026, doi: 10.1016/j.measurement.2025.119870.

[4] C. Lin, Y. Wang, B. Gong, H. Liu, and H. Liu, “Roadside LiDAR Deployment Optimization for Vehicle-to-Infrastructure Cooperative Perception in Urban Occlusion Environments,” *IEEE Trans. Instrum. Meas.*, vol. 74, pp. 1–14, 2025, doi: 10.1109/TIM.2025.3558231.

[5] H. Yu *et al.*, “DAIR-V2X: A Large-Scale Dataset for Vehicle-Infrastructure Cooperative 3D Object Detection,” in *Proc. IEEE/CVF Conf. Comput. Vis. Pattern Recognit. (CVPR)*, New Orleans, LA, USA: IEEE, Jun. 2022, pp. 21329–21338. doi: 10.1109/CVPR52688.2022.02067.

[6] X. Ye *et al.*, “Rope3D: The Roadside Perception Dataset for Autonomous Driving and Monocular 3D Object Detection Task,” in *Proc. IEEE/CVF Conf. Comput. Vis. Pattern Recognit. (CVPR)*, New Orleans, LA, USA: IEEE, Jun. 2022, pp. 21309–21318. doi: 10.1109/CVPR52688.2022.02065.

[7] W. Zimmer, C. Creß, H. T. Nguyen, and A. C. Knoll, “TUMTraf Intersection Dataset: All You Need for Urban 3D Camera-LiDAR Roadside Perception,” in *Proc. IEEE Int. Conf. Intell. Transp. Syst. (ITSC)*, Bilbao, Spain: IEEE, Sep. 2023, pp. 1030–1037. doi: 10.1109/ITSC57777.2023.10422289.

[8] A. Geiger, P. Lenz, C. Stiller, and R. Urtasun, “Vision meets robotics: The KITTI dataset,” *Int. J. Robot. Res.*, vol. 32, no. 11, pp. 1231–1237, Sep. 2013, doi: 10.1177/0278364913491297.

[9] H. Caesar *et al.*, “nuScenes: A Multimodal Dataset for Autonomous Driving,” in *Proc. IEEE/CVF Conf. Comput. Vis. Pattern Recognit. (CVPR)*, Seattle, WA, USA: IEEE, Jun. 2020, pp. 11618–11628. doi: 10.1109/CVPR42600.2020.01164.

[10] P. Sun *et al.*, “Scalability in Perception for Autonomous Driving: Waymo Open Dataset,” in *Proc. IEEE/CVF Conf. Comput. Vis. Pattern Recognit. (CVPR)*, Seattle, WA, USA: IEEE, Jun. 2020, pp. 2443–2451. doi: 10.1109/CVPR42600.2020.00252.

[11] A. Chen, P. J. Jin, and T. T. Zhang, “LiDAR Vehicle Trajectory Reconstruction with Arterial Shockwave Detection and Space–Time Analysis,” *Transp. Res. Rec.*, vol. 2680, no. 1, pp. 497–518, 2026.

[12] X. Zhou, C. Wang, Q. Xie, and T. Qiu, “V2I-Coop: Accurate Object Detection for Connected Automated Vehicles at Accident Black Spots With V2I Cross-Modality Cooperation,” *IEEE Trans. Mob. Comput.*, vol. 24, no. 3, pp. 2043–2055, Mar. 2025, doi: 10.1109/TMC.2024.3486758.

[13] J. Zhang *et al.*, “Roadside LiDAR-based scene understanding toward intelligent traffic perception: A comprehensive review,” *ISPRS J. Photogramm. Remote Sens.*, vol. 233, pp. 69–88, 2026, doi: 10.1016/j.isprsjprs.2026.01.012.

[14] Y. Li *et al.*, “V2X-Sim: Multi-Agent Collaborative Perception Dataset and Benchmark for Autonomous Driving,” *IEEE Robot. Autom. Lett.*, vol. 7, no. 4, pp. 10914–10921, Oct. 2022, doi: 10.1109/LRA.2022.3192802.

[15] S. Manivasagam *et al.*, “LiDARsim: Realistic LiDAR Simulation by Leveraging the Real World,” in *Proc. IEEE/CVF Conf. Comput. Vis. Pattern Recognit. (CVPR)*, Jun. 2020, pp. 11164–11173. doi: 10.1109/CVPR42600.2020.01118.

[16] A. Carballo *et al.*, “LIBRE: The Multiple 3D LiDAR Dataset,” in *Proc. IEEE Intell. Veh. Symp. (IV)*, Oct. 2020, pp. 1094–1101. doi: 10.1109/IV47402.2020.9304681.

[17] J. Schulte-Tigges *et al.*, “Benchmarking of Various LiDAR Sensors for Use in Self-Driving Vehicles in Real-World Environments,” *Sensors*, vol. 22, no. 19, p. 7146, Sep. 2022, doi: 10.3390/s22197146.

[18] G. Zhuang, Z. Bing, X. Yao, Y. Huang, K. Huang, and A. Knoll, “Toward Intelligent Sensing: Optimizing Lidar Beam Distribution for Autonomous Driving,” *IEEE Trans. Intell. Transp. Syst.*, vol. 24, no. 8, pp. 8386–8392, Aug. 2023, doi: 10.1109/TITS.2023.3267961.

[19] Y. He, P. Cao, D. Suo, and X. Liu, “A Joint Optimization of Beam Distribution and Deployment for Roadside LiDAR Systems to Maximize Vehicle Perception,” *IEEE Trans. Intell. Veh.*, vol. 10, no. 2, pp. 1300–1314, Feb. 2025, doi: 10.1109/TIV.2024.3426524.

[20] Y. He, D. Suo, P. Cao, and X. Liu, “Optimizing roadside LiDAR beam distribution to enhance vehicle detection performance considering dynamic vehicle occlusion effects,” *Transp. Res. Part C Emerg. Technol.*, vol. 179, p. 105268, Oct. 2025, doi: 10.1016/j.trc.2025.105268.

[21] M. Tang *et al.*, “A Multi-Scene Roadside LiDAR Benchmark towards Digital Twins of Road Intersections,” *ISPRS Ann. Photogramm. Remote Sens. Spat. Inf. Sci.*, vol. X-4–2024, pp. 341–348, Oct. 2024, doi: 10.5194/isprs-annals-X-4-2024-341-2024.

[22] Q. Pu, Y. Zhu, J. Wang, H. Yang, K. Xie, and S. Cui, "Drone Data Analytics for Measuring Traffic Metrics at Intersections in High-Density Areas," *Transp. Res. Rec.*, p. 03611981241311566, 2025.
[23] E. Li, S. Wang, C. Li, D. Li, X. Wu, and Q. Hao, "SUSTechPOINTS: A Portable 3D Point Cloud Interactive Annotation Platform System," in *Proc. IEEE Intell. Veh. Symp. (IV)*, 2020, pp. 1108–1115.
[24] S. Cui, P. Cao, Y. Wang, D. Suo, and X. Liu, "Analytical Models for Assessing LiDAR Perception Performance for Vehicles with and without Occlusion," *IEEE Trans. Veh. Technol.*, pp. 1–16, 2025, doi: 10.1109/TVT.2025.3617472.
[25] T. Yin, X. Zhou, and P. Krahenbuhl, "Center-based 3D Object Detection and Tracking," in *Proc. IEEE/CVF Conf. Comput. Vis. Pattern Recognit. (CVPR)*, Nashville, TN, USA: IEEE, Jun. 2021, pp. 11779–11788. doi: 10.1109/CVPR46437.2021.01161.
[26] Y. Yan, Y. Mao, and B. Li, "SECOND: Sparsely Embedded Convolutional Detection," *Sensors*, vol. 18, no. 10, Oct. 2018, doi: 10.3390/s18103337.
[27] T. Zhu, W. Li, and Y. Feng, "Integrated optimization for vehicle trajectory reconstruction under cooperative perception environment," *Transp. Res. Part C Emerg. Technol.*, vol. 184, p. 105522, Mar. 2026, doi: 10.1016/j.trc.2026.105522.

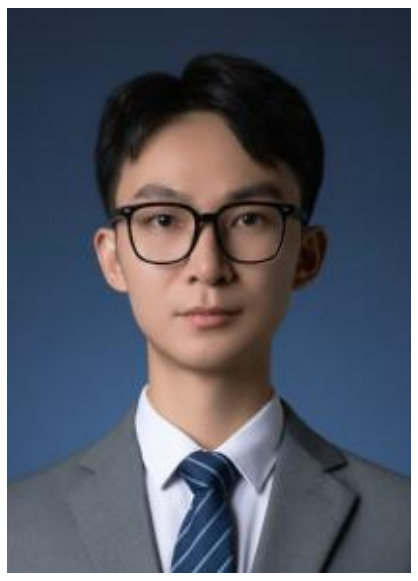

**Shunlai Cui** received his B.S. degree from Inner Mongolia University, Hohhot, China, in 2022, and his M.S. degree from Southwest Jiaotong University, Chengdu, China, in 2025. He is currently pursuing his Ph.D. degree at Old Dominion University, Norfolk, VA, USA. His research interests include LiDAR sensor perception and V2X safety assessment.

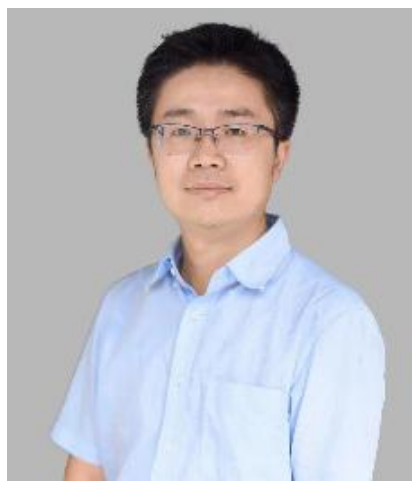

**Peng Cao** earned his B.S. degree in Industrial Engineering from the University of Electronic Science and Technology of China, Chengdu, China in 2009, followed by an M.S. in Management Science and Engineering from Tsinghua University, Beijing, China in 2011, and a Ph.D. in Civil Engineering from Nagoya University, Nagoya, Japan in 2014. He is now an Associate Professor at the School of Transportation and Logistics, Southwest Jiaotong University, Chengdu, China. His research focuses on Traffic Data Collection and Traffic Flow Optimization in the context of Connected Autonomous Vehicles.

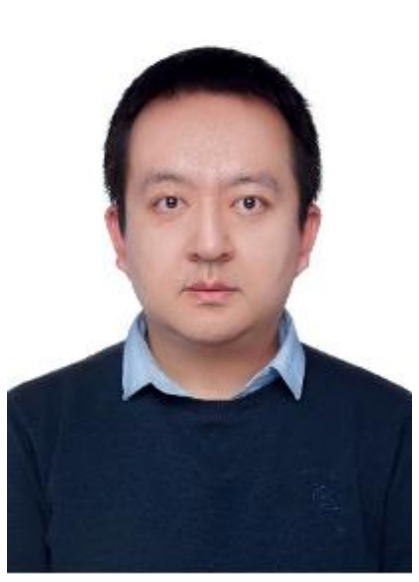

**Yuan Zhu** is currently a professor of Transportation Institute, Inner Mongolia University (IMU). He received his B.S. degree in software engineering from Jilin University in 2009, M.S. degrees in Computer Science and Transportation Planning and Engineering in 2011 and 2013 from Polytechnic Institute of NYU, and Ph.D. degree in Transportation Planning and Engineering and from New York University in 2018. He was enrolled in "Grassland Talent" Program by Inner Mongolia Autonomous Region. His research interests include emergency evacuation, multi-scopic simulations and traffic data analytics.

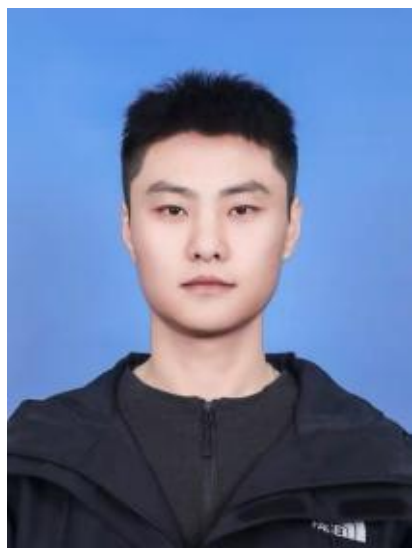

**Yongjiang He** received the B.S. and M.S. degrees in transportation engineering from Shandong University of Science and Technology, Qingdao, China, in 2018 and 2021, respectively, and the Ph.D. degree in transportation engineering from Southwest Jiaotong University, Chengdu, China, in December 2025. He is currently an Assistant Professor with the Tangshan Research Institute, Southwest Jiaotong University, Tangshan, China. His research focuses on intelligent transportation systems (ITS), particularly roadside sensing technologies, sensor deployment optimization, and perception performance modeling in vehicle–infrastructure cooperative environments. His work aims to develop efficient and reliable sensing and deployment strategies to enhance traffic safety, perception robustness, and large-scale intelligent transportation infrastructure implementation.

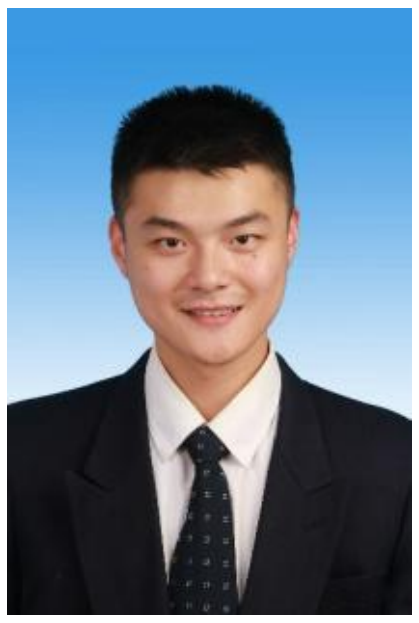

**Jiacheng Yin** received the Ph.D. degree from Southwest Jiaotong University, in 2024. He is currently a Lecturer with the School of Automobile and Transportation, Xihua University. His research interests include traffic flow modeling and analysis in complex environments, low-altitude traffic situation awareness, and representation.

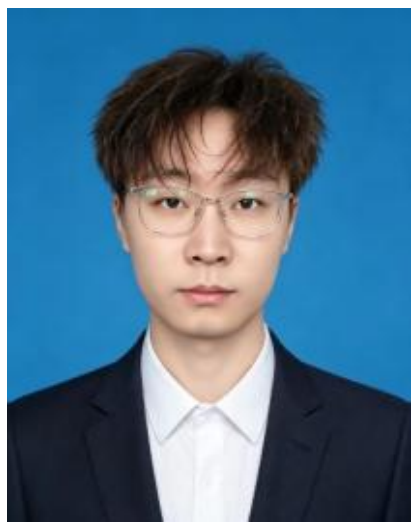

**Xiao Huo** earned his B.S. degree in Transportation from Chongqing Jiaotong University, Chongqing, China, in 2024. He is currently pursuing his M.S. degree at Southwest Jiaotong University, Chengdu, China. His research interests include LiDAR placement optimization and V2X perception.

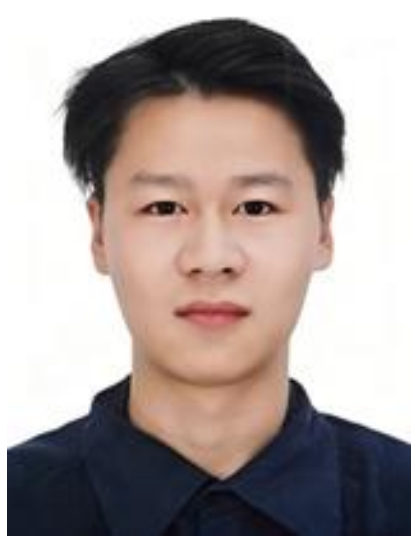

**Gang Cao** earned his B.S. degree in Logistics Engineering from Southwest Jiaotong University, Chengdu, China, in 2025. He is currently pursuing his M.S. degree in Transportation Engineering at Southwest Jiaotong University, Chengdu, China. His research interests include artificial intelligence, object detection, and object tracking.

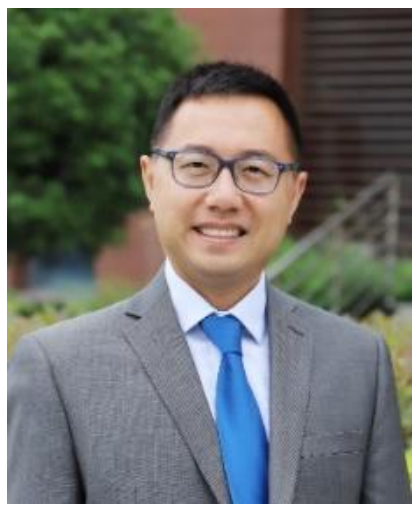

**Xiaobo Liu** earned his Ph.D. degree from the New Jersey Institute of Technology, Newark, New Jersey, in 2004. He is currently a professor with the School of Transportation and Logistics, Southwest Jiaotong University, Chengdu, China. His research focuses on the direction of transportation system analysis under connected vehicle/autonomous vehicle environment, and intelligent logistics analysis. He received the George Krambles Transportation Scholarship, 2003; Most Outstanding Student Paper Award by the Institute of Transportation Engineers Metropolitan Section of New York and New Jersey, 2004; and Stella Dafermos Best Paper Award by the Transportation Research Board Transportation Network Modeling Committee, 2018.